\documentclass{article}

\usepackage{microtype}
\usepackage{graphicx}
\usepackage{subcaption}
\usepackage{booktabs} 
\usepackage{pifont}
\usepackage{multirow}
\usepackage{tabularx}
\usepackage[most]{tcolorbox}
\usepackage{xcolor}
\usepackage{amsmath}
\usepackage{amssymb}
\usepackage{amsfonts}
\usepackage{mathrsfs}
\usepackage{bm}
\usepackage{array}
\usepackage{makecell}
\usepackage{longtable}
\usepackage{colortbl}
\usepackage{listings}
\usepackage{enumitem}
\usepackage{nicefrac}
\usepackage{url}
\usepackage{xspace}
\usepackage{relsize}
\usepackage{placeins}

\usepackage{arydshln}

\usepackage{hyperref}

\usepackage[preprint]{icml2026}

\usepackage{amsmath}
\usepackage{amssymb}
\usepackage{mathtools}
\usepackage{amsthm}

\usepackage[capitalize,noabbrev]{cleveref}

\theoremstyle{plain}

\theoremstyle{definition}

\theoremstyle{remark}

\usepackage[textsize=tiny]{todonotes}
\providecommand{\Require}{\REQUIRE}
\providecommand{\Ensure}{\ENSURE}
\providecommand{\State}{\STATE}
\providecommand{\Statex}{\STATE}
\makeatletter
\renewcommand{\Notice@String}{}
\renewcommand{\printAffiliationsAndNotice}[1]{\global\icml@noticeprintedtrue%
  \stepcounter{@affiliationcounter}%
  {\let\thefootnote\relax\footnotetext{\hspace*{-\footnotesep}\ificmlshowauthors #1\fi%
      \forloop{@affilnum}{1}{\value{@affilnum} < \value{@affiliationcounter}}{
        \textsuperscript{\arabic{@affilnum}}\ifcsname @affilname\the@affilnum\endcsname%
          \csname @affilname\the@affilnum\endcsname%
        \else
          {\bf AUTHORERR: Missing \textbackslash{}icmlaffiliation.}
        \fi
      }.%
    }
  }
}
\makeatother
\definecolor{darkgreen}{RGB}{0, 100, 0}
\definecolor{codegreen}{rgb}{0,0.3,0.6}
\definecolor{codegray}{rgb}{0.5,0.5,0.5}
\setlist[itemize]{noitemsep, topsep=0pt}
\newcommand{\gain}[1]{\hspace{2pt}\textcolor{red}{\scriptsize(+{#1})}}
\newcommand{\equalcontrib}{\texorpdfstring{\textsuperscript{*}}{*}}
\icmltitlerunning{AAPA: Adversarially Anchored Preference Alignment for Post-Training of Large Language Models}

\begin{document}

\twocolumn[
  \icmltitle{AAPA: Adversarially Anchored Preference Alignment \\ for Post-Training of Large Language Models}

  \begin{icmlauthorlist}
    \icmlauthor{Faqiang Qian\equalcontrib}{sensetime}
    \icmlauthor{Kang An\equalcontrib}{sensetime}
    \icmlauthor{Weikun Zhang\equalcontrib}{sensetime}
    \icmlauthor{Ziliang Wang}{sensetime}
    \icmlauthor{Xuhui Zheng}{sensetime}
    \par\vspace{0.15em}
    \icmlauthor{Liangjian Wen}{swufe}
    \icmlauthor{Yong Dai}{xhumanoid}
    \icmlauthor{Mengya Gao}{sensetime}
    \icmlauthor{Yichao Wu}{sensetime}
  \end{icmlauthorlist}

  {\centering
  \texttt{\{qianfaqiang,gaomengya,wuyichao\}@senseauto.com}\par
  \vspace{0.08in}
  }

  \icmlaffiliation{sensetime}{SenseTime}
  \icmlaffiliation{swufe}{Southwest University of Finance and Economics}
  \icmlaffiliation{xhumanoid}{X-Humanoid}

  \icmlkeywords{Large Language Models, Alignment, Preference Learning, Post-Training}

  \vskip 0.3in
]



\printAffiliationsAndNotice{\textsuperscript{*}Equal contribution. }  

\begin{abstract}
Post-training alignment of large language models often combines supervised fine-tuning (SFT) on expert demonstrations with reinforcement learning (RL) from preference or verifiable feedback. SFT provides a useful behavioral anchor but can overfit to static demonstrations, whereas RL encourages exploration but may drift from expert behavior or exploit imperfect rewards. We propose \textbf{AAPA} (\emph{Adversarially Anchored Preference Alignment}), a plug-in framework that augments existing post-training objectives with a sentence-level adversarial anchoring signal. AAPA compares policy rollouts with offline, pre-collected expert responses using a fixed lightweight discriminator, and therefore requires neither online teacher inference nor discriminator co-training during policy optimization. The same anchoring term can be added to SFT, GRPO, and CHORD while preserving their original training pipelines.
Experiments on instruction-following benchmarks show that AAPA consistently improves the corresponding base objectives across model scales. In particular, the staged AAPA configuration improves over a strong GRPO baseline by 5.77\% on \texttt{Qwen3-0.6B} and 3.75\% on \texttt{Qwen3-4B}. Further analyses on response length, log-probability distributions, and discriminator variants suggest that adversarial anchoring provides a stable semantic grounding signal for preference optimization.
Code is available at \url{https://github.com/IsFaqq/AAPA}.

\end{abstract}

\section{Introduction}

\begin{figure}[t]
    \centering
    \includegraphics[width=0.4\textwidth]{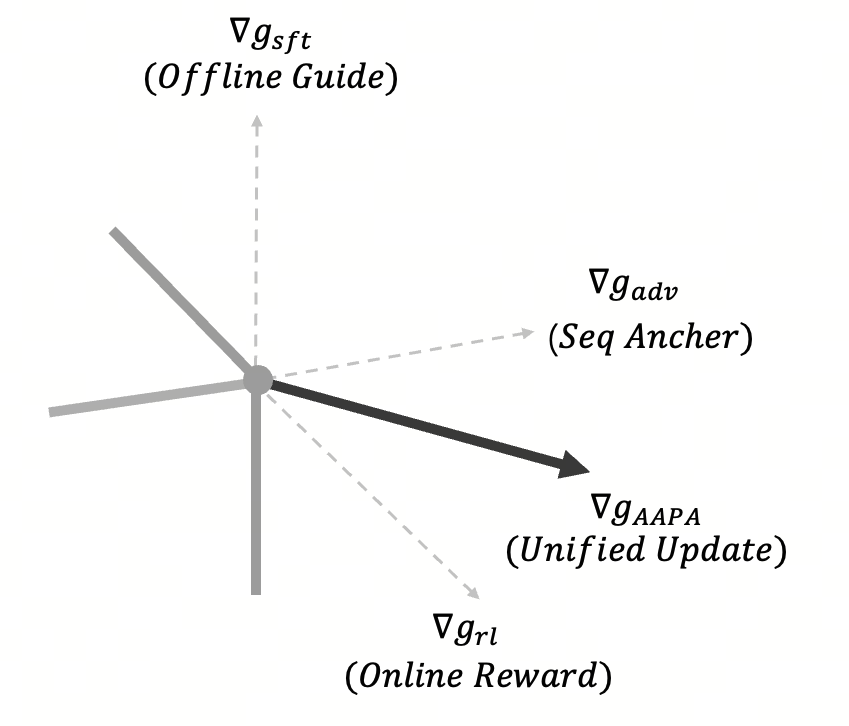}
    \caption{Unified gradient interpretation of AAPA.
AAPA augments existing post-training objectives with an adversarial anchoring term computed from sentence-level comparisons between policy rollouts and offline expert responses. Under this view, A-SFT and A-GRPO are supervised or preference-based updates with the same fine-grained anchor.}
    \label{fig:maing}
    \vspace{-4pt}
\end{figure}

Large Language Models (LLMs) have demonstrated strong capabilities in reasoning, generation, and human–computer interaction~\cite{first1, first2}. As these models become increasingly capable and autonomous, aligning their behavior with human intent and values has become a central challenge~\cite{first3}. Post-training alignment addresses this challenge by shaping pretrained models into helpful, harmless, and reliable systems, and is now a foundational component of contemporary LLM development~\cite{first4}.

\begin{figure*}[t]
    \centering
    \includegraphics[width=0.9\textwidth]{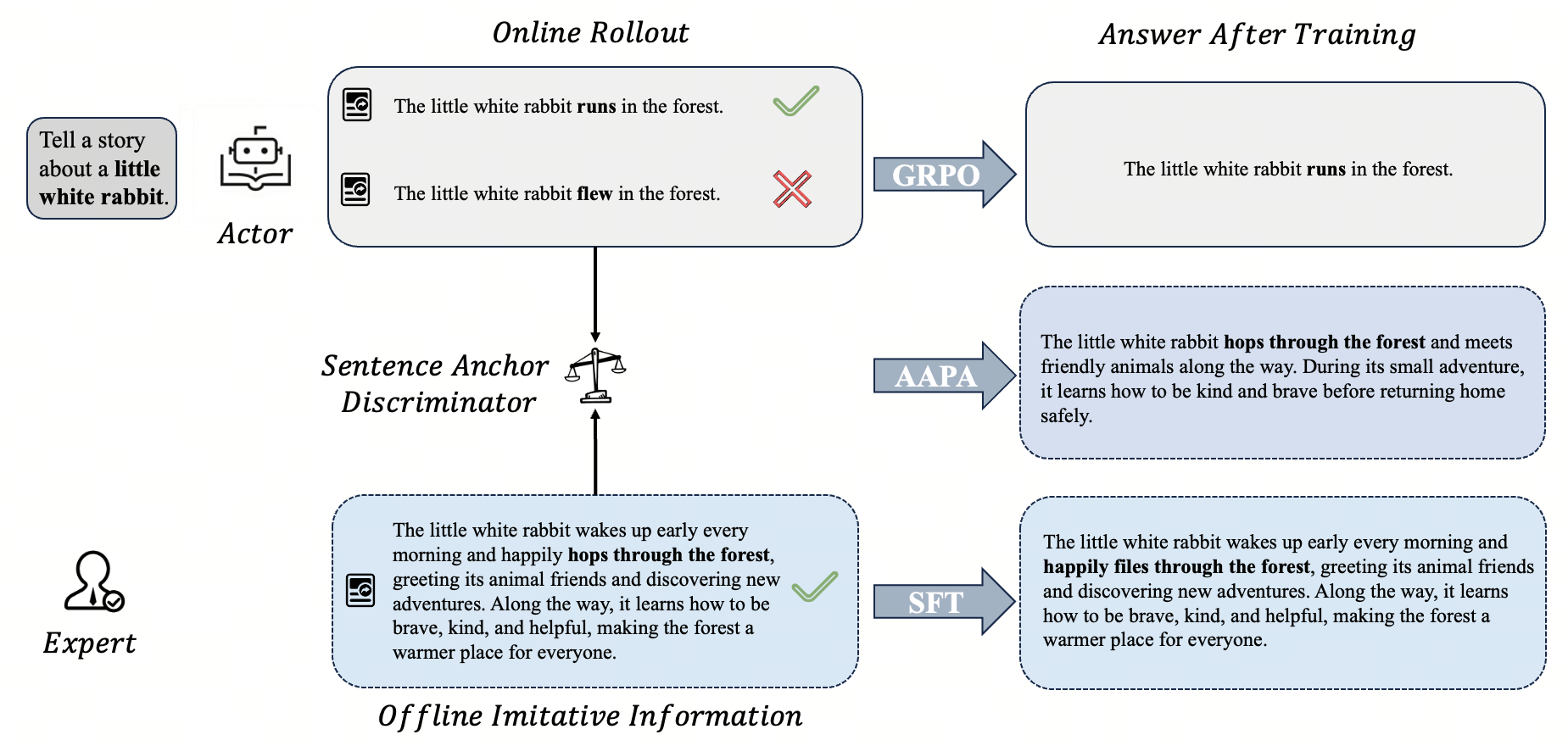}
    \caption{Overview of AAPA as a plug-in sentence-level adversarial anchoring framework.
For the same input prompt, AAPA decomposes policy rollouts and offline, pre-collected expert responses into sentences, and uses a fixed discriminator to compare their semantic relations at this finer granularity. The resulting sentence-level anchoring signal can be added to different post-training objectives, providing distributional grounding while preserving the original SFT, GRPO, or CHORD training pipeline.}
    \label{fig:mainimage}
\end{figure*}

\begin{table*}[t]
    \centering
    \renewcommand{\arraystretch}{1.0}
    \resizebox{0.9\textwidth}{!}{
    \begin{tabular}{|l|c|c|c|c|c|c|}
        \hline
        \multirow{2}{*}{\textbf{Feature}} 
        & \multicolumn{3}{c|}{\textbf{Non-Adversarial}} 
        & \multicolumn{3}{c|}{\textbf{Adversarial (Ours)}} \\
        \cline{2-7}
        & \textbf{SFT} & \textbf{GRPO} & \textbf{CHORD} 
        & \textbf{A-SFT} & \textbf{A-GRPO} & \textbf{A-CHORD} \\
        \hline
        Offline Question     
        & \textcolor{darkgreen}{\checkmark} 
        & \textcolor{darkgreen}{\checkmark} 
        & \textcolor{darkgreen}{\checkmark} 
        & \textcolor{darkgreen}{\checkmark} 
        & \textcolor{darkgreen}{\checkmark}
        & \textcolor{darkgreen}{\checkmark} \\
        \hline
        Offline Response     
        & \textcolor{darkgreen}{\checkmark} 
        & \textcolor{red}{\ding{55}}        
        & \textcolor{darkgreen}{\checkmark} 
        & \textcolor{darkgreen}{\checkmark} 
        & \textcolor{darkgreen}{\checkmark} 
        & \textcolor{darkgreen}{\checkmark} \\
        \hline
        Online Exploration       
        & \textcolor{red}{\ding{55}}        
        & \textcolor{darkgreen}{\checkmark} 
        & \textcolor{darkgreen}{\checkmark} 
        & \textcolor{darkgreen}{\checkmark} 
        & \textcolor{darkgreen}{\checkmark}
        & \textcolor{darkgreen}{\checkmark} \\
        \hline
        Reward Constraints   
        & \textcolor{red}{\ding{55}}        
        & \textcolor{darkgreen}{\checkmark} 
        & \textcolor{darkgreen}{\checkmark} 
        & \textcolor{red}{\ding{55}}        
        & \textcolor{darkgreen}{\checkmark}
        & \textcolor{darkgreen}{\checkmark} \\
        \hline
        Token Level          
        & \textcolor{darkgreen}{\checkmark} 
        & \textcolor{darkgreen}{\checkmark} 
        & \textcolor{darkgreen}{\checkmark} 
        & \textcolor{darkgreen}{\checkmark} 
        & \textcolor{darkgreen}{\checkmark} 
        & \textcolor{darkgreen}{\checkmark} \\
        \hline
        Sentence Level       
        & \textcolor{red}{\ding{55}}        
        & \textcolor{red}{\ding{55}}        
        & \textcolor{red}{\ding{55}}        
        & \textcolor{darkgreen}{\checkmark} 
        & \textcolor{darkgreen}{\checkmark}
        & \textcolor{darkgreen}{\checkmark} \\
        \hline
    \end{tabular}
    }
    \vspace{2mm}
    \caption{Comparison of training paradigms in terms of data usage, exploration strategy, reward constraints, and loss granularity.}
    \label{tab:comparison_transposed}
\end{table*}

In this work, we revisit post-training alignment as a problem of preference learning. Human intent is communicated to models through two complementary forms of preference data. Demonstrated preferences consist of expert-provided examples of desired behavior, from which models learn via supervised fine-tuning (SFT), acquiring foundational capabilities, stylistic conventions, and safety constraints—what we refer to as \emph{imitative information}~\cite{second1}. In contrast, comparative preferences arise from human judgments over alternative model outputs~\cite{second2}, and are typically leveraged through reinforcement learning methods such as Direct Preference Optimization (DPO)~\cite{dpo} and Group Relative Policy Optimization (GRPO)~\cite{grpo} to further refine model behavior. Although both forms of preference data aim to shape aligned behavior, existing objectives often use them through different optimization signals, leaving a gap between expert grounding and policy exploration.

A central challenge in such alignment pipelines is the distributional mismatch between static expert data and the model's dynamically evolving policy. SFT offers a useful behavioral anchor but optimizes only on offline demonstrations, while RL improves the policy under its own rollout distribution and may receive sparse or imperfect feedback. This mismatch gives rise to two tightly coupled learning failures.

\begin{itemize}
    \item \textbf{Offline Learning (Imitative Brittleness):} 
    Supervised fine-tuning optimizes on a fixed expert distribution, providing a strong initial knowledge base but risking overfitting to static demonstrations~\cite{third0}. As the policy evolves away from the expert data~\cite{third00}, the imitative knowledge learned via SFT can become brittle beyond the training distribution.
    
    \item \textbf{Online Learning (Ungrounded Exploration):} 
    Online reinforcement learning improves the policy through exploration beyond expert data, but this exploration is weakly grounded in the dense knowledge encoded in demonstrations. As a result, learning can become inefficient and unstable, and remains susceptible to policy drift and reward-hacking behaviors~\cite{third1, third2}.
\end{itemize}

These failures suggest that expert demonstrations should not only initialize a policy, but also provide a persistent grounding signal as the policy explores. To this end, we propose \textbf{AAPA} (\emph{Adversarially Anchored Preference Alignment}), a plug-in framework that augments post-training objectives with sentence-level adversarial anchoring. Given offline, pre-collected expert responses, AAPA uses a fixed lightweight discriminator to compare expert and policy responses at the sentence level, producing a semantic anchoring signal that can regularize SFT-, GRPO-, or CHORD-style training without requiring online teacher inference or discriminator co-training. Table~\ref{tab:comparison_transposed} summarizes how these instantiations differ in data usage, exploration strategy, reward constraints, and loss granularity. Our main contributions are as follows:
\begin{itemize}
    \item \textbf{Problem Formulation and Framework.} We formalize post-training alignment as a distributional mismatch between offline expert demonstrations and evolving policy rollouts, and propose AAPA, a plug-in sentence-level adversarial anchoring framework to bridge this gap (Figure~\ref{fig:mainimage}).
    
    \item \textbf{Plug-in Instantiations.} 
    We instantiate the same anchoring signal with SFT, GRPO, and CHORD, yielding A-SFT, A-GRPO, and A-CHORD while preserving the original training pipelines.
    
    \item \textbf{Empirical Validation.} We show that staged AAPA substantially outperforms a strong GRPO baseline, with up to 5.77\% and 3.75\% absolute improvements on \texttt{Qwen3-0.6B} and \texttt{Qwen3-4B}. We further analyze response length and log-probability distributions, and extend evaluation to mathematical reasoning tasks.
\end{itemize}

\section{Related Work}

\noindent\textbf{Supervised Fine-Tuning.} Supervised fine-tuning is a cornerstone of post-training for LLMs due to its simplicity and effectiveness. However, maximum-likelihood training can overfit demonstrations, resulting in overconfidence and degraded generalization. To address these limitations, researchers have explored regularization strategies~\cite{regular1, regular2, regular3} and high-quality data synthesis methods~\cite{synth0,synth00,synth1,synth2,lin2024idgen}. More recently, reinforcement learning (RL) has been integrated into the post-training pipeline, forming the dominant ``SFT-then-RL'' paradigm used by state-of-the-art LLMs~\cite{qwen3,deepseek,gemma,gpt4, cheng2023adversarial}.

\noindent\textbf{Reinforcement Learning.} Reinforcement learning aligns model outputs with human preferences or verifiable feedback. Recent work applies RL to domains with verifiable solutions, such as mathematics and programming, yielding notable gains under reinforcement learning with verifiable rewards (RLVR)~\cite{tulu3, deepseekr1, imbrlvr}. However, RL samples from the model's own output distribution, making exploration inefficient when the model lacks knowledge for difficult instances. While curriculum learning~\cite{autoglm} and dynamic sampling~\cite{dapo} can mitigate this issue, external expert knowledge provides a more direct guide.

\noindent\textbf{Incorporating External Knowledge into Reinforcement Learning.} Incorporating external knowledge can improve training efficiency~\cite{sftmemory,qsft,chord}, especially when repeated RL failures indicate that a problem exceeds the model's current ability. Prior work uses high-quality external responses either in rollouts~\cite{offpolicydata1} or as exemplars for policy guidance~\cite{offpolicydata2,offpolicydata3}. The former breaks the on-policy nature of RL, while the latter introduces state inconsistencies between exploration and learning. Other methods~\cite{srft, chord} add cross-entropy loss to RL, but token-level likelihood maximization can reduce entropy and hamper exploration. Unlike prior approaches, AAPA keeps expert responses offline and uses a fixed discriminator to provide sentence-level semantic anchoring between policy outputs and expert demonstrations, following the broader use of adversarial mechanisms for bridging distribution gaps~\cite{apo}.

\section{Methodology}

\subsection{Theoretical Foundation: The Alignment Duality and Dilemma}
The objective of LLM alignment is to find a policy $\pi_\theta$ that reflects human intent, conveyed through \textit{Imitative Information} from expert demonstrations $D_{\text{SFT}}$ and \textit{Preference Information} from feedback $D_{\text{PREF}}$. An aligned policy should benefit from both sources: expert demonstrations provide a behavioral reference, while preference or verifiable feedback encourages improvement under the policy's own rollout distribution.

\paragraph{Optimality Condition for Alignment.}
We formalize this intuition as a constrained optimization problem:
\[
\begin{aligned}
\pi_{\text{aligned}}
&=\arg\max_{\pi_\theta}
\mathbb{E}_{x,y\sim\pi_\theta}[R_\psi(y|x)] \\
&\quad \text{s.t.}\;
\mathcal{D}(\pi_\theta\|\pi_{\text{exp}})\le\epsilon .
\end{aligned}
\]

Here $R_\psi$ denotes the reward or verification signal from Preference Information, $\pi_{\text{exp}}$ denotes the expert behavior induced by offline demonstrations, and $\mathcal{D}$ is a divergence measure. The constraint requires the learned policy to remain grounded in expert behavior while optimizing for preference-based rewards. The sequential ``SFT-then-RL'' pipeline can be viewed as one practical strategy for this objective, but it may become suboptimal when the policy distribution drifts far from the offline expert distribution during RL.

\subsection{The AAPA Framework: Adversarial Bridge for the Distributional Gap}
AAPA operationalizes this objective by adding a sentence-level adversarial anchoring term to existing post-training losses. Rather than requiring online teacher interaction, AAPA assumes a set of offline, pre-collected expert responses paired with the training prompts. These responses provide a semantic reference against which policy rollouts can be compared during training.

\paragraph{The Adversarial Anchoring Signal.} We employ a fixed sentence-level discriminator, $D_\phi$, to estimate semantic consistency between a policy rollout and its corresponding expert response. Given a prompt $x$, a policy response $y_s \sim \pi_{\theta}(\cdot|x)$, and an offline expert response $y_t$ associated with $x$, the discriminator outputs a score $D_{\phi}(x, y_t, y_s) \in [-1, 1]$. Higher scores indicate that the policy response is semantically closer to the expert response at the sentence level. The corresponding auxiliary objective is:
\[
\begin{aligned}
\mathcal{L}_{\text{ADV}}(\theta; \phi)
= -\mathbb{E}_{x \sim p(x),\, y_s \sim \pi_{\theta}(\cdot|x)}
\big[D_\phi(x, y_t, y_s)\big].
\end{aligned}
\]
Since $y_s$ is sampled text, this term is used as an auxiliary policy-level signal rather than as direct back-propagation through discrete tokens. In SFT-style training, it augments the supervised objective through rollout-based scoring; in RL-style training, it is incorporated as an additional reward-like anchoring term. In both cases, the resulting update encourages the policy to improve while remaining semantically grounded in the offline expert response.

\subsection{Gradient Formulations of Alignment Components}
AAPA provides a common anchoring term that can be plugged into different post-training objectives. We describe its instantiations for SFT and GRPO below. Appendix~\ref{app:pref_algos} provides a discussion of DPO and KTO.

\paragraph{Adversarial Supervised Fine-Tuning (A-SFT).}
\subparagraph{Objective:} To preserve the imitation ability of SFT while adding a sentence-level semantic anchor to expert responses.
\subparagraph{Methodology:} Standard SFT is driven by the negative log-likelihood loss, $\mathcal{L}_{\text{SFT}}(\theta) = - \mathbb{E}_{(x,y) \sim D_{\text{SFT}}} [\log \pi_\theta(y|x)]$. A-SFT augments this objective with the adversarial anchoring term:
\begin{equation*}
\mathcal{L}_{\text{A-SFT}}(\theta; \phi)
= \mathcal{L}_{\text{SFT}}(\theta)
+ \lambda_{\text{adv}} \mathcal{L}_{\text{ADV}}(\theta; \phi).
\end{equation*}
The coefficient $\lambda_{\text{adv}}$ controls the strength of the anchoring signal relative to the supervised imitation loss.

\paragraph{Adversarial Group Relative Policy Optimization (A-GRPO).}
\subparagraph{Objective:} To guide RL exploration toward higher-reward regions while grounding policy updates in offline expert responses.
\subparagraph{Methodology:} This stage utilizes an online RL algorithm, GRPO, which is driven by a reward model $RM_\psi$ trained on $D_{\text{PREF}}$. For a given prompt, $G$ outputs are sampled from a behavior policy $\pi_{\text{old}}$, and their advantages $\hat{A}_i$ are calculated based on normalized rewards. The GRPO update is derived from its PPO-style loss function, $\mathcal{L}_{\text{GRPO}}$, which combines a clipped surrogate objective $\mathcal{J}_{\text{CLIP}}$ and a KL penalty. This loss is defined as:

\[
\begin{aligned}
\mathcal{L}_{\text{GRPO}}(\theta,\pi_{\text{old}},\pi_{\text{ref}})
&= - \mathcal{J}_{\text{CLIP}}(\theta,\pi_{\text{old}}) \\
&\quad + \mathbb{E}_{q\sim D_{\text{PREF}}}
\big[ \beta_D\, \Delta_{\text{KL}}(q) \big],
\end{aligned}
\]
where $\Delta_{\text{KL}}(q)=D_{\mathrm{KL}}(\pi_\theta(\cdot|q)\|\pi_{\text{ref}}(\cdot|q))$.

A-GRPO augments this objective with the same adversarial anchoring term:
\begin{equation*}
\begin{aligned}
\mathcal{L}_{\text{A-GRPO}}(\theta; \phi)
&= \mathcal{L}_{\text{GRPO}}(\theta,\pi_{\text{old}},\pi_{\text{ref}}) \\
&\quad + \lambda_{\text{adv}} \mathcal{L}_{\text{ADV}}(\theta; \phi).
\end{aligned}
\end{equation*}
Equivalently, the discriminator score can be treated as an auxiliary reward-like signal when computing GRPO advantages.

\subsection{Optional Mixed-Objective Instantiation}
Beyond adding AAPA to individual stages, the same anchoring term can also be used in a mixed-objective training setup. Similar hybrid objectives such as CHORD~\cite{chord} can be augmented analogously by adding the same adversarial anchoring term while preserving their original dynamic weighting mechanisms. We present this formulation as one possible instantiation rather than as the central claim of AAPA. We provide further discussion in Appendix~\ref{app:sft_as_pref} and Appendix~\ref{app:achord_formulation}.

\paragraph{Methodology:} We construct mixed batches from $D_{\text{SFT}}$ and $D_{\text{PREF}}$ and optimize:
\[
\begin{aligned}
\mathcal{L}_{\text{Mixed}}(\theta; \phi)
&= \alpha \cdot \mathcal{L}_{\text{A-SFT}}(\theta; \phi) \\
&\quad + (1-\alpha) \cdot \mathcal{L}_{\text{A-GRPO}}(\theta; \phi),
\end{aligned}
\]
where $\alpha$ controls the relative weight between supervised imitation and preference optimization.

\paragraph{Gradient Interpretation.} The gradient of this mixed objective is a weighted sum of the gradients from the A-SFT and A-GRPO components, computed on their respective data slices within the mixed batch:
\[
\begin{aligned}
\mathbf{g}_{\text{Mixed}}
&= \nabla_\theta \mathcal{L}_{\text{Mixed}} \\
&= \alpha \cdot \mathbf{g}_{\text{A-SFT}}|_{\text{SFT}}
+ (1-\alpha) \cdot \mathbf{g}_{\text{A-GRPO}}|_{\text{PREF}} .
\end{aligned}
\]

By substituting the definitions of the component gradients, as illustrated in Figure~\ref{fig:maing}, we observe that the final update vector is a coordinated combination of three fundamental learning signals. In particular, the adversarial gradient can be further decomposed into two learning signals:

\[
\begin{aligned}
\mathbf{g}_{\text{Mixed}}
&= \underbrace{
\alpha \nabla_\theta \mathcal{L}_{\text{SFT}}
}_{\substack{\text{Imitation Signal}\\\text{(from SFT data)}}}
\\
&\quad + \underbrace{
(1-\alpha)\nabla_\theta \mathcal{L}_{\text{GRPO}}
}_{\substack{\text{Preference Signal}\\\text{(from RL data)}}} \\
&\quad + \underbrace{
\lambda_{\text{adv}}\Delta_{\text{ADV}}
}_{\substack{\text{Global Distributional}\\\text{Regularization}}} .
\end{aligned}
\]
where $\Delta_{\text{ADV}}=\alpha\nabla_\theta \mathcal{L}_{\text{ADV}}|_{\text{SFT}}+(1-\alpha)\nabla_\theta \mathcal{L}_{\text{ADV}}|_{\text{PREF}}$.

This interpretation shows how AAPA can provide distributional grounding in both supervised and preference-based updates. In our experiments, we evaluate separated, staged, and mixed-objective variants to distinguish the effect of the adversarial anchor from the choice of training schedule.

\section{Experiments}
In this section, we evaluate the performance of AAPA on instruction-following tasks and conduct a comprehensive analysis of the model's output behavior. Additional experiments on mathematical reasoning and reward modeling are provided in Appendix~\ref{app:supplyablation}.

\subsection{Experimental Setup}
\noindent\textbf{Datasets.} We use three instruction-following datasets—AutoIF~\cite{autoif}, IFevallike~\cite{ifevallike}, and IFBench~\cite{ifbench}—as our training data, with sizes of 61k, 56k, and 38k examples, respectively. The 61k and 56k examples come from the original AutoIF and IFevallike datasets, while the 38k ground-truth responses (GT) in IFBench are obtained from the original 95k examples. Each example consists of a prompt, a verification function, and a corresponding GT that passes the verification function. Examples of the training data, validation functions, and response data can be found in Appendix~\ref{app:exampletrain}.

\noindent\textbf{Teacher and Sentence-level Discriminator.}
Teacher responses are generated before policy optimization by running \texttt{Qwen3-235B-Instruct-2507} on prompts from the training datasets, and are kept fixed throughout training. We additionally include an ablation study that adopts \texttt{Qwen3-32B} as the teacher model; detailed results are provided in Appendix~\ref{app:supplyablation}. Note that teacher-generated responses are not guaranteed to satisfy the verification function.

For the sentence-level discriminator, our framework does not rely on a specialized architecture and can be instantiated with standard sentence-level semantic discriminators. In the main experiments, we adopt and modify POLAR~\cite{polar} as a universal discriminator. POLAR is a reference-based reward model that evaluates whether a response and its reference are likely to originate from the same policy, producing a corresponding similarity score. The specific modifications that adapt POLAR for use as a universal sentence-level discriminator are detailed in Appendix~\ref{app:adversarial_impl}. 

We further conduct supplementary robustness checks, including a BGE-M3 discriminator variant and two length-based anchoring variants, to separate semantic anchoring from simple length effects.

\begin{table*}[t]
\renewcommand\arraystretch{1.1} 
\centering
\begin{tabular}{lcccccc}
\toprule
\multirow{2}{*}{\textbf{\textit{Method}}}  & \multicolumn{6}{c}{Benchmarks} \\
\cmidrule(lr){2-7}
 & Avg & I-Following & English & Coding & Mathematics & Chinese \\
\midrule

\textbf{\textit{Qwen3-0.6B}} & 40.81 & 56.62 & 32.61 & 32.63 & 42.29 & 43.25 \\

\textbf{\textit{SFT}}      
& 38.48 & 59.97 & 25.20 & 29.43 & 39.56 & 44.33 \\

\textbf{\textit{A-SFT}}    
& 39.83 & 60.72\gain{0.75} & 26.67 & \textbf{33.93} & 40.14 & 44.10 \\

\textbf{\textit{GRPO}}     
& 40.93 & 74.09 & 32.39 & 17.58 & 41.49 & 43.11 \\

\textbf{\textit{A-GRPO}}   
& 41.99 & 79.58\gain{5.49} & 31.36 & 19.79 & 41.04 & 43.96 \\

\textbf{\textit{SFT$\rightarrow$ GRPO}}   
& 42.56 & 77.40 & 35.09 & 21.60 & 39.57 & 44.38 \\

\textbf{\textit{SFT$\rightarrow$ A-GRPO}} 
& \textbf{43.61} & 79.57\gain{2.17} & 32.71 & 29.00 & 39.95 & 44.10 \\

\textbf{\textit{A-SFT$\rightarrow$ A-GRPO}} 
& 43.31 & \textbf{79.86}\gain{2.46} & \textbf{36.48} & 33.72 & \textbf{43.72} & \textbf{44.47} \\
\hdashline
\textbf{\textit{CHORD}}    
& 41.19 & 73.52 & 29.92 & 26.09 & 38.52 & 44.90 \\

\textbf{\textit{A-CHORD}}  
& \textbf{42.09} & \textbf{75.25}\gain{1.73} & \textbf{32.74} & 23.89 & 39.45 & \textbf{45.14} \\

\midrule

\textbf{\textit{Qwen3-4B}} & 62.52 & 80.10 & 45.42 & 70.31 & 59.47 & 67.36 \\

\textbf{\textit{SFT}}      
& 60.80 & 80.60 & 41.13 & 68.62 & 56.92 & 68.51 \\

\textbf{\textit{A-SFT}}    
& 60.67 & 82.47\gain{1.87} & 38.78 & 70.34 & 56.65 & 68.07 \\

\textbf{\textit{GRPO}}     
& 68.11 & 86.90 & 53.42 & 71.43 & \textbf{66.52} & 70.42 \\

\textbf{\textit{A-GRPO}}   
& \textbf{68.39} & 88.17\gain{1.27} & \textbf{54.29} & 73.15 & 65.12 & 69.89 \\

\textbf{\textit{SFT$\rightarrow$ GRPO}}   
& 65.75 & 89.75 & 49.04 & 71.54 & 59.40 & 70.53 \\

\textbf{\textit{SFT$\rightarrow$ A-GRPO}} 
& 68.11 & 90.45\gain{0.70} & 48.25 & 72.06 & 60.53 & 70.93 \\

\textbf{\textit{A-SFT$\rightarrow$ A-GRPO}} 
& 67.93 & \textbf{90.65}\gain{0.90} & 51.16 & \textbf{73.26} & 63.65 & \textbf{71.45} \\
\hdashline
\textbf{\textit{CHORD}}    
& \textbf{69.30} & 86.97 & 54.86 & \textbf{73.76} & 67.67 & 71.30 \\

\textbf{\textit{A-CHORD}}  
& 69.27 & \textbf{88.27}\gain{1.30} & \textbf{55.46} & 70.32 & \textbf{68.14} & \textbf{71.66} \\

\bottomrule
\end{tabular}
\caption{Comparative results of different training methods across multiple benchmarks (IFEval, MulitIF, MMLU, MMLU-Pro, GPQA-Diamond, HumanEval, MBPP, GSM8K, MATH-500, TheoremQA, CMMLU, CEval) for models with \texttt{Qwen3-0.6B} and \texttt{Qwen3-4B}.Bold numbers indicate the best results achieved under either staged or unified training. Red annotations indicate absolute improvements on instruction-following performance over the corresponding non-adversarial baselines.}
\label{tab:mainstage}
\end{table*}

\begin{figure}[t]
    \centering
    \includegraphics[width=0.48\textwidth]{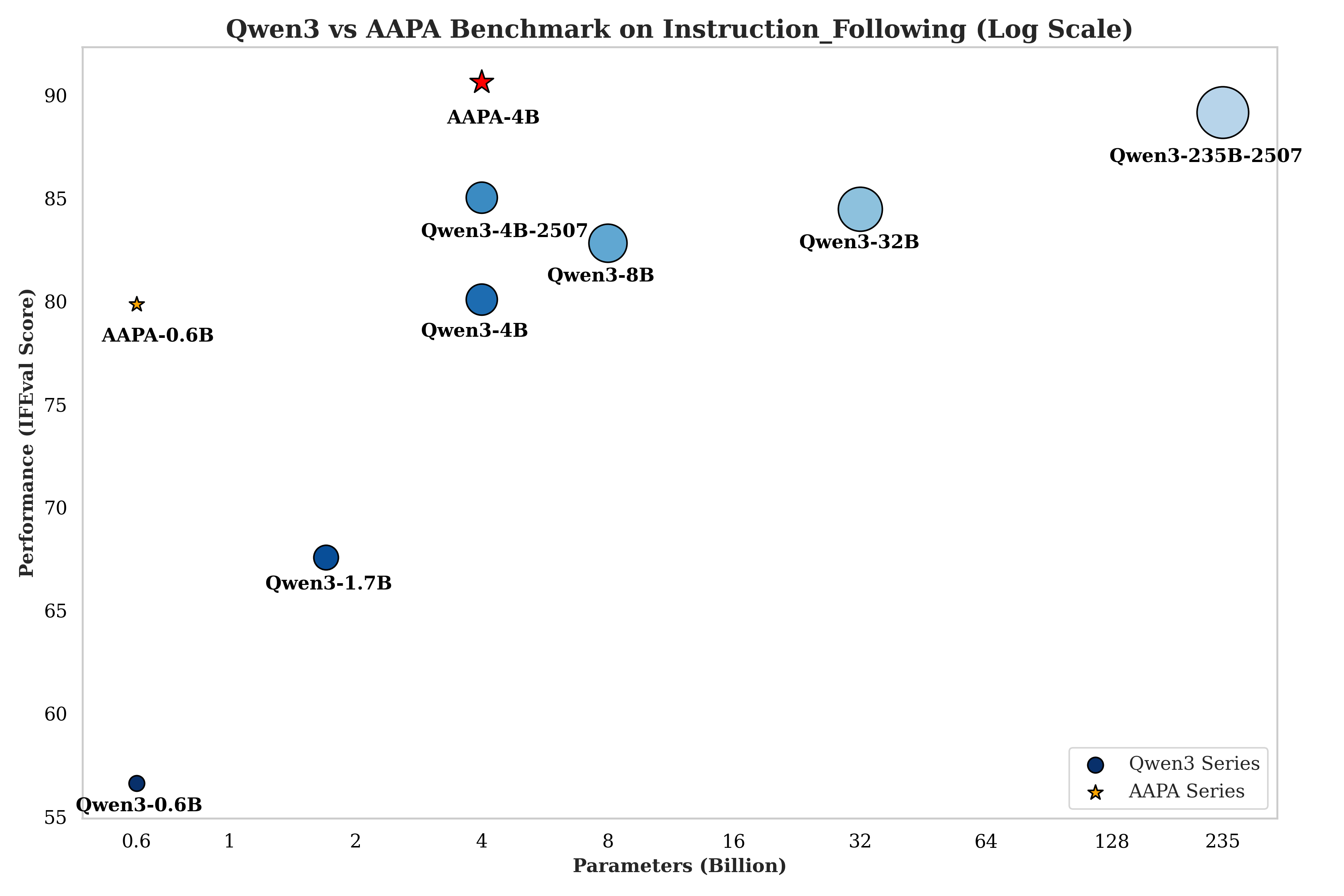}
    \caption{
    AAPA performance on the instruction-following benchmark (average of IFEval and MultiIF).
    Notably, \texttt{AAPA-0.6B} performs on par with the much larger \texttt{Qwen3-4B} model,
    and \texttt{AAPA-4B} exceeds the teacher baseline on this benchmark,
    \texttt{Qwen3-235B-A22B-Instruct-2507}.
    }
    \label{fig:mainresult}
    \vspace{-6pt}
\end{figure}

\noindent\textbf{Benchmarks.} To verify the effectiveness of our method, we evaluate it on two common instruction-following benchmarks and ten general-purpose tasks, including IFEval~\cite{ifeval}, MultiIF~\cite{multiif}, general English (MMLU~\cite{mmlu}, MMLU-Pro~\cite{mmlupro}, GPQA-Diamond~\cite{GPQA}), coding (Humaneval~\cite{humaneval}, Mbpp~\cite{mbpp}), mathematics (GSM8K~\cite{gsm8k}, Math-500~\cite{math500}, TheoremQA~\cite{theoremqa}), and Chinese (CMMLU~\cite{cmmlu}, CEval~\cite{ceval}). For MultiIF, we evaluate only the single-turn multilingual instruction-following ability. We report the average performance per domain in the main tables, with detailed results for each dataset provided in Appendix~\ref{app:detailresult}.

\noindent\textbf{Setting.}
We evaluate whether sentence-level adversarial anchoring remains useful across model scales and training schedules. Specifically, we incorporate AAPA into supervised fine-tuning (SFT), reinforcement learning (RL), and mixed-objective settings, training Qwen3 models at two scales: 0.6B and 4B. Appendix~\ref{app:additional_robustness} further reports 8B A-GRPO scaling and 0.6B random-seed robustness checks.

In the SFT setting, the training objective is the standard cross-entropy loss between model predictions and ground-truth responses. For RL, we directly adopt GRPO as the optimization algorithm. Under verifiable reward settings, the model receives a reward of 1 if its response passes the verification function, and 0 otherwise. In addition, we include CHORD~\cite{chord}, a hybrid method that jointly performs supervised fine-tuning (SFT) and reinforcement learning (RL), as a baseline to further validate the generality of our adversarial loss formulation.

We evaluate the effect of integrating AAPA at different stages of training. \textbf{A-SFT} augments the cross-entropy objective with the adversarial loss, \textbf{A-GRPO} incorporates the adversarial signal into the GRPO objective, and \textbf{A-CHORD} extends CHORD by adding the adversarial loss to its hybrid optimization objective.

All experiments are conducted using MS-Swift~\cite{swift}, VeRL~\cite{VeRL}, and Trinity-RFT~\cite{trinity-rft}. Detailed experimental configurations are provided in Appendix~\ref{app:experimental_settings}.

\subsection{Main Result}
\noindent\textbf{Objective Metrics and Experimental Analysis.}
Figure~\ref{fig:mainresult} presents the staged results and compares them against the Qwen3 series models on instruction-following tasks. The full set of evaluation metrics is provided in Appendix~\ref{app:detailresult}. 
As shown in Table~\ref{tab:mainstage}, several key findings emerge from the staged training setup. 
First, in the offline supervised fine-tuning (SFT) phase, A-SFT slightly mitigates the degradation in general capabilities and achieves larger gains on instruction-following benchmarks compared to standard SFT.
Second, during the online training phase, A-GRPO achieves performance comparable to the staged SFT\textrightarrow GRPO approach on task-specific metrics. One possible explanation is that A-GRPO continuously anchors student rollouts to offline teacher responses during training, partially serving the role of an implicit SFT warm-up. This effect is particularly evident in the 4B model, where A-GRPO mitigates the degradation of general capabilities typically introduced by SFT. 
Furthermore, AAPA improves the corresponding base objectives across staged and mixed-objective settings, suggesting that the adversarial anchor is useful beyond a single training schedule.
Moreover, by enhancing the model's instruction-following capability, the adversarial loss improves instruction comprehension and yields stronger performance across several capability categories compared to the base model.

\begin{figure}[t]
    \centering
    \begin{minipage}[b]{0.48\columnwidth}
        \centering
        \includegraphics[width=\columnwidth]{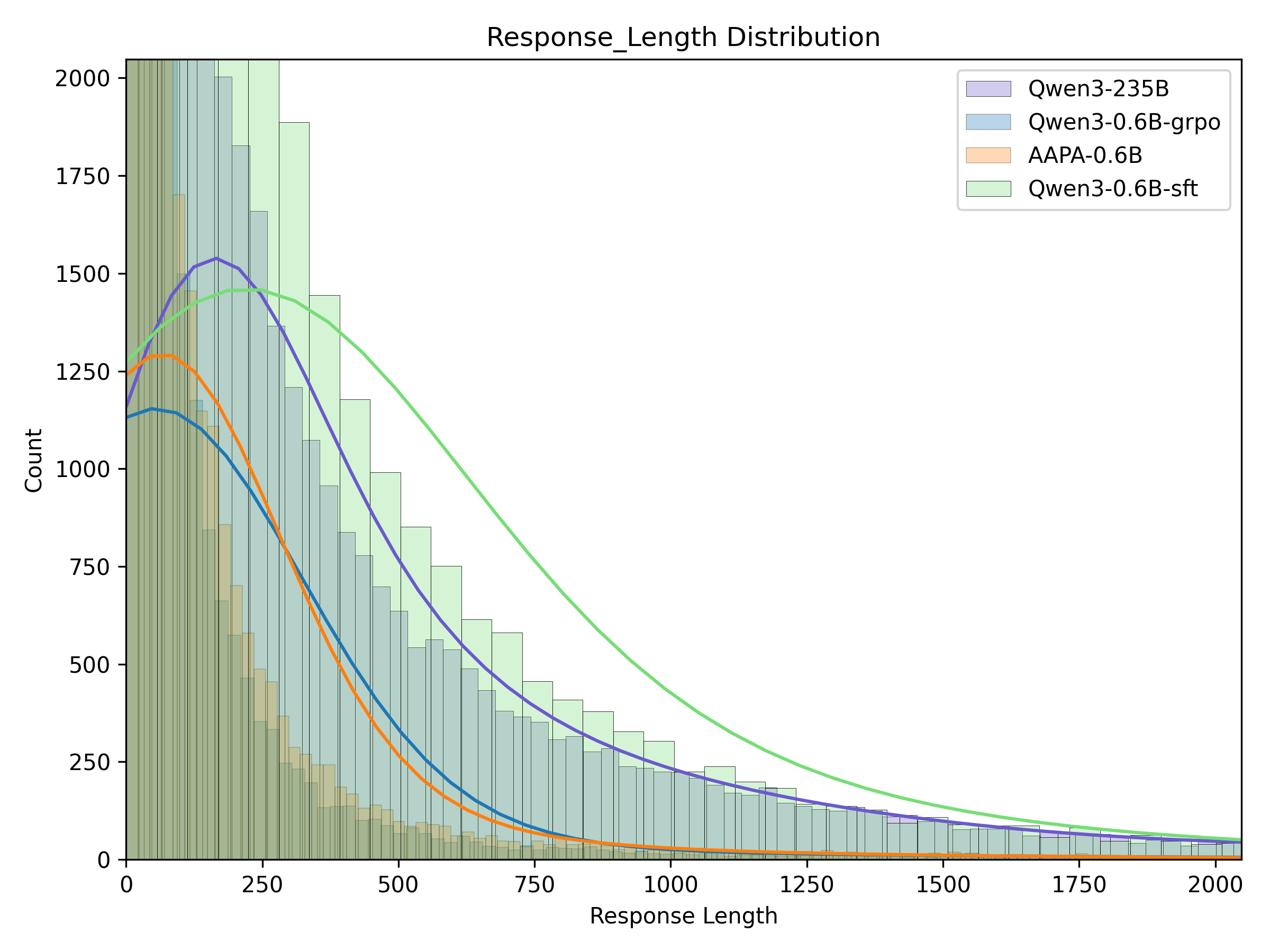}
        \subcaption{Response length distributions.}
        \label{fig:response_dif}
    \end{minipage}
    \hfill
    \begin{minipage}[b]{0.48\columnwidth}
        \centering
        \includegraphics[width=\columnwidth]{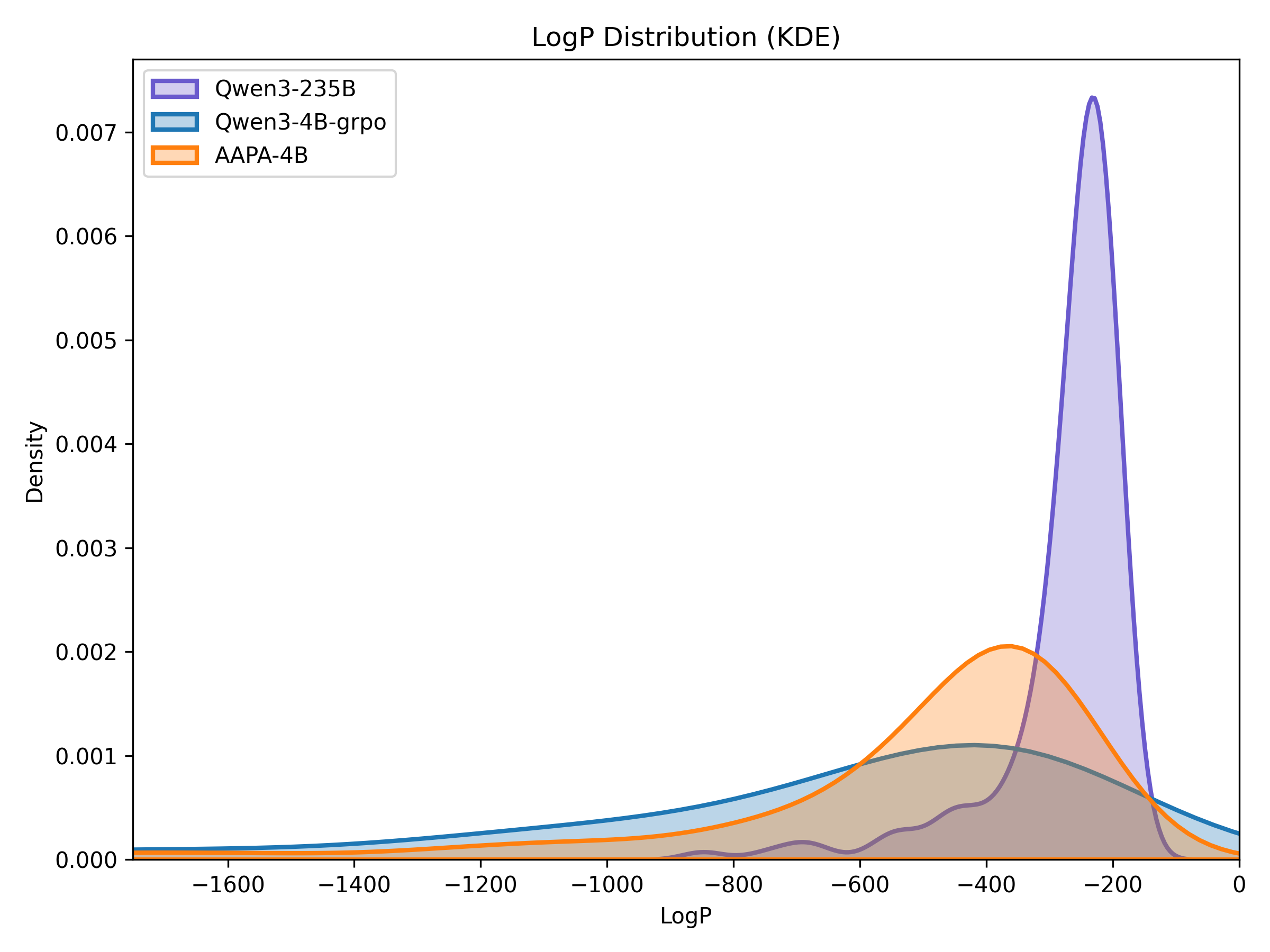}
        \subcaption{Log-probability difference KDE.}
        \label{fig:LogpDif}
    \end{minipage}
    \caption{
    Distributional comparison between student and teacher models.
    AAPA produces response length and log-probability distributions that more closely match the teacher than GRPO.
    }
    \label{fig:distribution_analysis}
    \vspace{-6pt}
\end{figure}

\noindent\textbf{Distributional Analysis.}
To assess whether AAPA provides a grounding signal that mitigates policy drift, we re-evaluate models trained with SFT, GRPO, and AAPA on the 38K instances from IFBench. Figure~\ref{fig:response_dif} reports the distribution of response lengths. Compared to GRPO, AAPA produces a response length distribution that more closely matches that of the teacher. This suggests that sentence-level anchoring can guide policy exploration toward the offline expert response distribution. A representative example is provided in Appendix~\ref{app:badcase}.

To further assess whether AAPA constrains the student toward the teacher distribution, we sample 500 responses and compute their total log-likelihood under three models: the teacher, a GRPO-trained student, and an AAPA-trained student. As shown in Figure~\ref{fig:LogpDif}, AAPA exhibits a noticeably narrower log-likelihood distribution with reduced divergence from the teacher. This concentration suggests stronger policy regularization and more effective knowledge transfer, indicating that AAPA maintains closer alignment with the teacher than GRPO during training.

\begin{figure}[t]
    \centering
    \begin{minipage}[b]{0.48\columnwidth}
        \centering
        \includegraphics[width=\columnwidth]{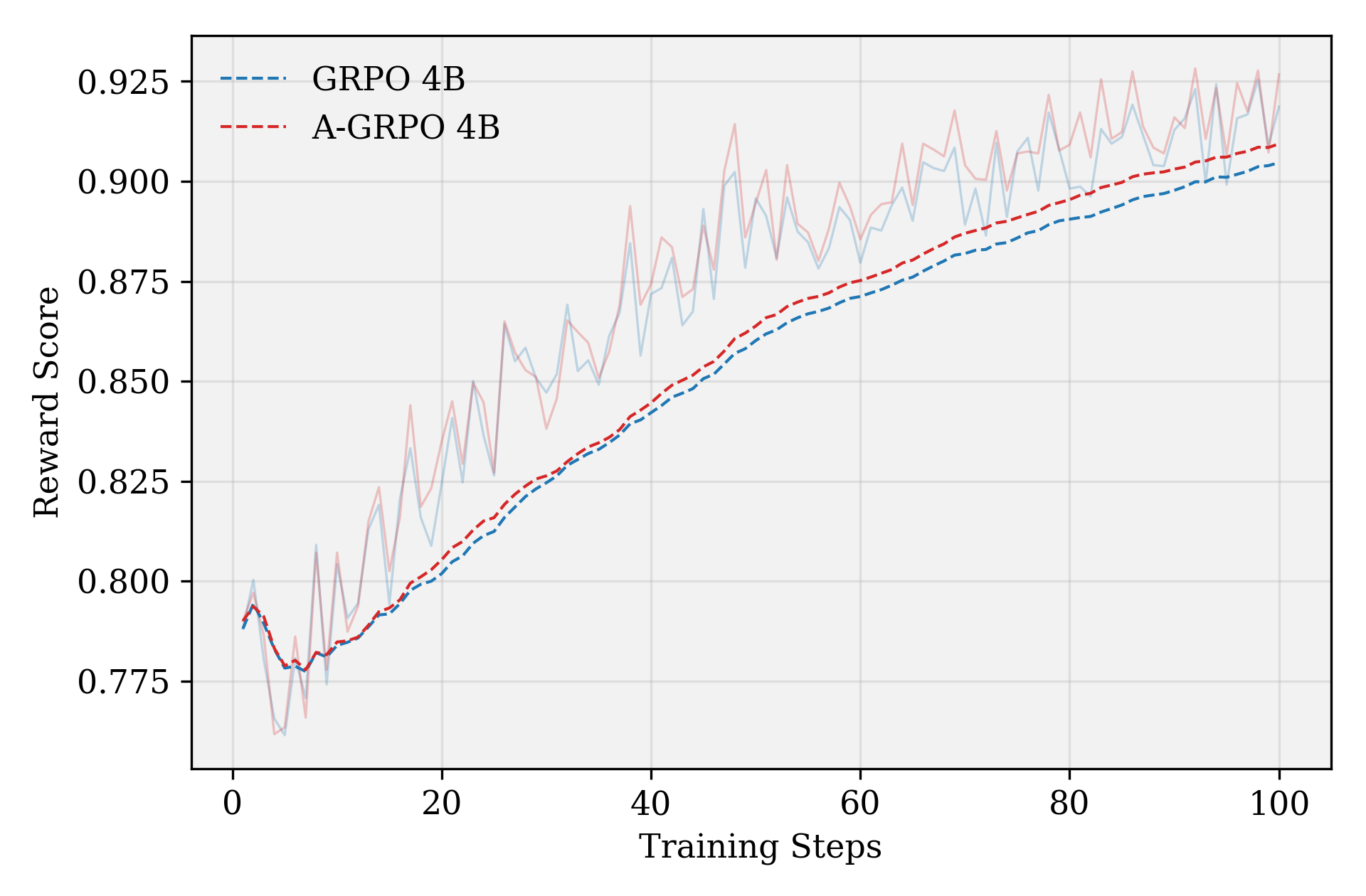}
        \subcaption{Reward acquisition.}
        \label{fig:reward}
    \end{minipage}
    \hfill
    \begin{minipage}[b]{0.48\columnwidth}
        \centering
        \includegraphics[width=\columnwidth]{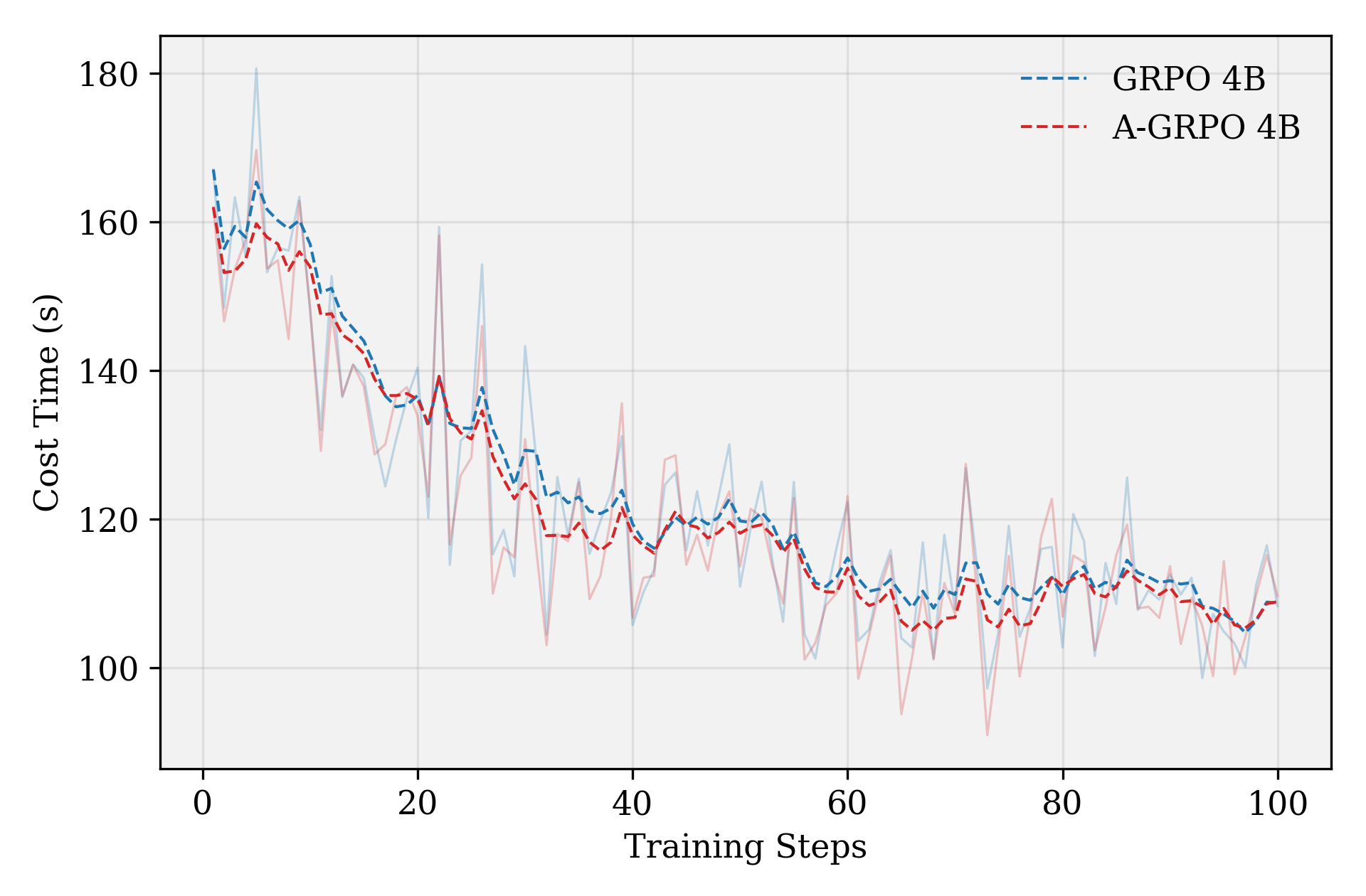}
        \subcaption{Actor update time.}
        \label{fig:actor_time}
    \end{minipage}
    \caption{
    Training dynamics of the 4B model in the first 100 steps.
    The adversarial-enhanced variant is shown in blue, while the GRPO baseline is shown in red.
    }
    \label{fig:training_dynamics}
    \vspace{-6pt}
\end{figure}

\noindent\textbf{Training Dynamics and Runtime.}
We further examine the early training behavior and computational overhead of AAPA. The additional loss term only introduces a scalar-weighted log-likelihood and therefore adds little cost to actor updates; as shown in Figure~\ref{fig:actor_time}, the actor update time remains essentially unchanged. Meanwhile, Figure~\ref{fig:reward} shows that incorporating adversarial gradients leads to higher reward acquisition within the first 100 training steps for the 4B model. In practice, discriminator evaluation can be served as a lightweight network call, and the main additional cost in RLVR depends on discriminator inference speed.

\begin{table*}[t]
    \renewcommand\arraystretch{1.1} 
    \centering
    \begin{tabular}{lcccccc}
        \hline
        \hline
        \toprule
        \multirow{2}{*}{\textbf{\textit{Method}}}  & \multicolumn{6}{c}{Benchmarks} \\
        \cmidrule(lr){2-7}
                & Avg & I-Following & English & Coding & Mathematics & Chinese  \\
        \midrule
        \textbf{\textit{RLVR}}        & 47.53 & 76.25 & \textbf{36.43} & 36.29 & 43.83 & 52.22 \\
        \textbf{\textit{RLVR+$\mathrm{\textit{D}_{output}}$}}  & 46.70 & 68.47 & 36.01 & 36.80 & \textbf{46.06} & 51.82 \\
        \textbf{\textit{RLVR+$\mathrm{\textit{A}_{coef}}$}}  & 47.68 & 73.09 & 35.96 & \textbf{38.94} & 45.09 & \textbf{52.51} \\
        \textbf{\textit{w $\mathrm{\textit{A}_{coef}}$}}   & \textbf{47.83} & \textbf{78.94}\gain{1.59} & 36.23 & 35.57 & 43.91 & 52.26 \\
        
        \bottomrule
    \end{tabular}
    \caption{RLVR denotes a setting that uses only verifiable rewards. The remaining three variants correspond to our ablation configurations.}
    \label{tab:ablation_method}

\end{table*}

\subsection{Ablation Study}
Unless otherwise specified, ablations are performed under the SFT$\rightarrow$A-GRPO training paradigm.

\paragraph{Ablation on AAPA Adversarial Loss Method.}
Specifically, we compared three settings in Table~\ref{tab:ablation_method}, where the base method employs RLVR (Reinforcement Learning with Verifiable Rewards)'s GRPO.
(i) directly adding the raw output of sentence-level discriminator  to the verifiable reward,
(ii) adding the transformed output to the verifiable reward,
(iii) using the transformed output as a separate adversarial loss term.
Among these settings, using it as an independent adversarial loss achieves the best performance, which aligns with our expectation that the anchor should act as a separate auxiliary signal rather than directly modifying the task reward.

\begin{figure}[t]
    \centering
    \begin{minipage}[b]{0.48\columnwidth}
        \centering
        \includegraphics[width=\columnwidth]{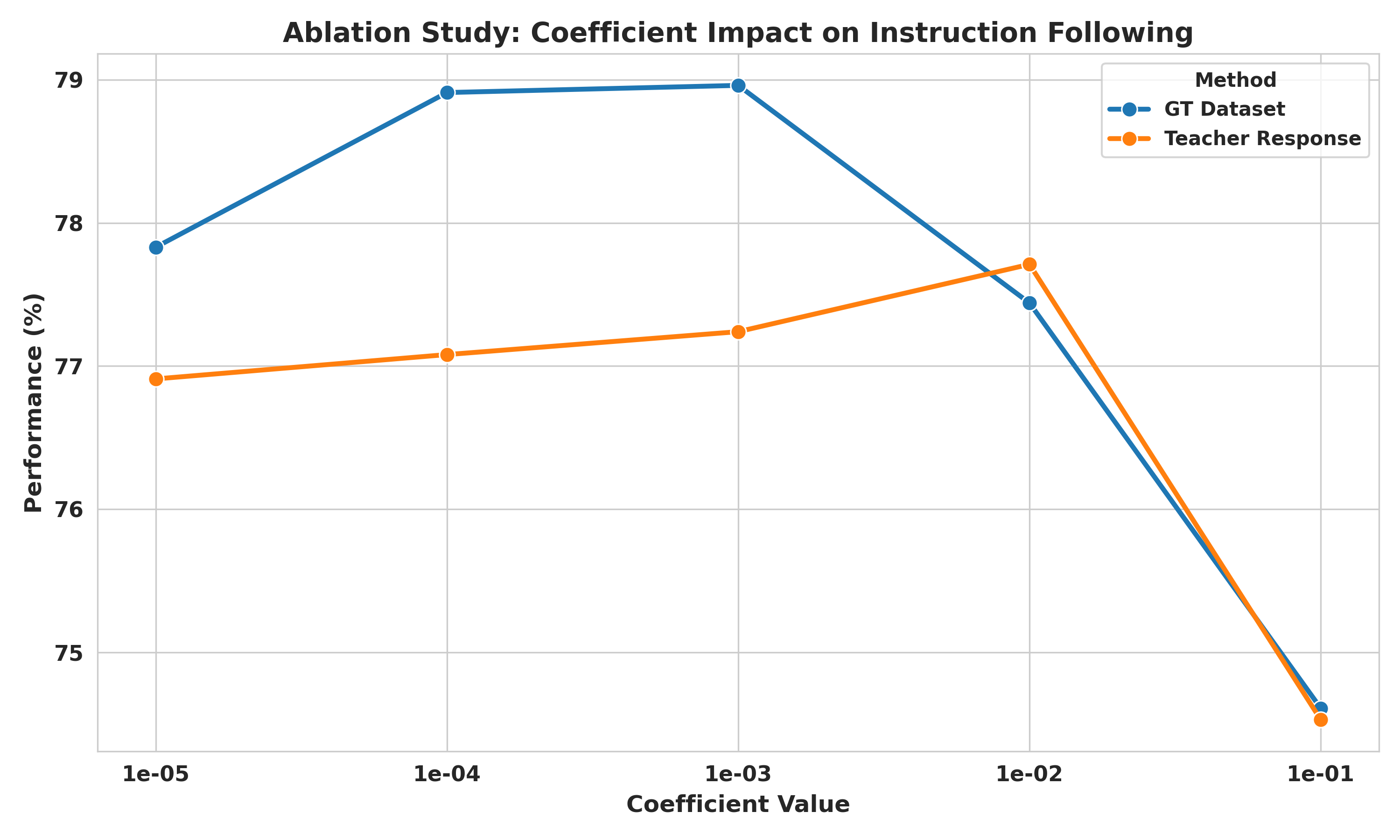}
        \subcaption{Instruction-following performance.}
        \label{fig:coefIF}
    \end{minipage}
    \hfill
    \begin{minipage}[b]{0.48\columnwidth}
        \centering
        \includegraphics[width=\columnwidth]{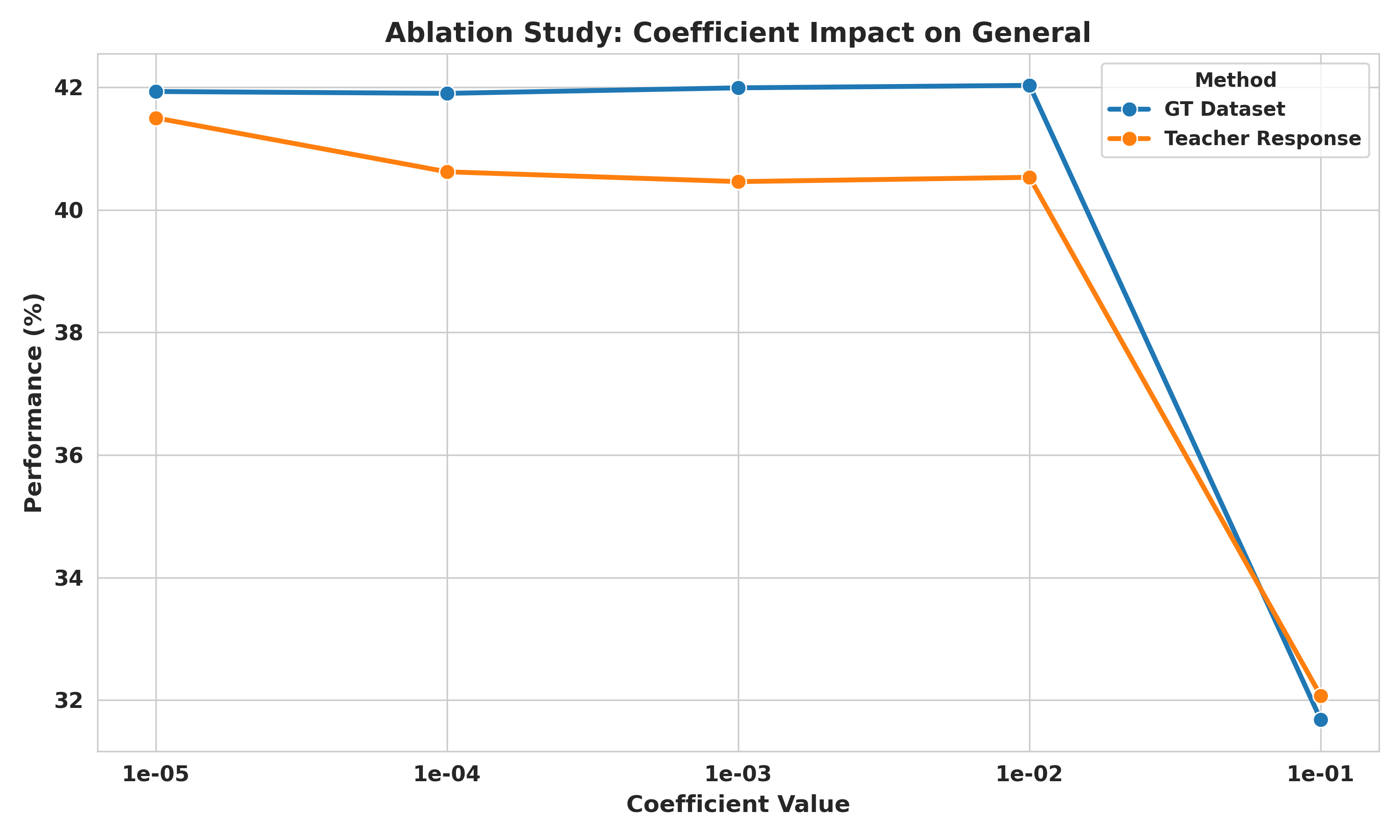}
        \subcaption{General-purpose performance.}
        \label{fig:coefgeneral}
    \end{minipage}
    \caption{
    Ablation study of discriminator coefficient $\lambda_{\mathrm{adv}}$ in AAPA.
    The results illustrate the trade-off between instruction-following ability and general capabilities.
    }
    \label{fig:coef_ablation}
    \vspace{-6pt}
\end{figure}

\paragraph{Ablation on the AAPA Adversarial Loss Coefficient \(\lambda_{\text{adv}}\).}
We conducted extensive ablation studies on the AAPA adversarial loss coefficient \(\lambda_{\text{adv}}\), 
varying it from \(0.1\) to \(1\times10^{-5}\), 
which spans the typical magnitude difference relative to the policy gradient loss. 
In all experiments, the coefficient of the KL loss was fixed at \(0.001\). 
Figure~\ref{fig:coefIF} and Figure~\ref{fig:coefgeneral} present all coefficient settings evaluated on both GT-based data and teacher-generated responses. We suggest setting the coefficient within the range of \(0.01\) to \(0.0001\), which is approximately one order of magnitude lower than the policy gradient loss, serving as an auxiliary signal. A value that is too high may dominate the optimization with the adversarial term and suppress the model's exploration, while a value that is too low may lead to insufficient performance gains.

\paragraph{Other Ablations.}
Additional ablations are provided in the appendix to test whether AAPA's gains depend on specific implementation choices. Appendix~\ref{app:experimental_settings} evaluates base-model initialization and ground-truth (GT) references, while Appendix~\ref{app:supplyablation} studies GT-reference discriminator coefficients, KL regularization, discriminator backbones, teacher scale, length-only anchoring, reward-model training, and mathematical reasoning. Appendix~\ref{app:additional_robustness} further reports random-seed robustness, 8B A-GRPO scaling, and a comparison with GKD, an on-policy distillation baseline that trains the student on self-generated responses with teacher feedback. These results show that teacher-feedback distillation and length matching do not replace online reward optimization with sentence-level anchoring, and that AAPA remains robust across discriminator, teacher scale, initialization, and training-schedule choices.

\section{Conclusion}
We propose \textbf{Adversarially Anchored Preference Alignment (AAPA)}, a unified framework that formulates post-training alignment as a constrained optimization problem to mitigate distributional mismatch. AAPA introduces an adversarial objective to enforce distributional consistency, enabling dense expert signals to dynamically regularize model exploration. This joint process anchors the policy to expert behavior while allowing preference learning to go beyond static expert distributions. Empirical results and analyses on instruction-following and mathematical reasoning tasks support our hypothesis that unifying imitation learning with preference-based learning yields a more effective and efficient alignment strategy. We believe AAPA provides a practical foundation for incorporating richer supervisory signals and guiding advanced AI systems.

\section*{Limitations}
Our experiments focus primarily on Qwen-family models and instruction-following tasks, with additional validation on mathematical reasoning and reward-model-based training. Although the results suggest that adversarial anchoring is broadly useful, further evaluation on more model families, multilingual settings, safety-critical domains, and human-preference datasets is needed. In addition, AAPA relies on the availability of a reasonably strong teacher response and a sentence-level discriminator; weak teachers or poorly calibrated discriminators may reduce the quality of the anchoring signal. Finally, while the extra training cost is modest in our setting, deployment-time choices such as discriminator serving, batching, and teacher-response generation may affect the overall efficiency of large-scale training pipelines.

\bibliography{main}
\bibliographystyle{icml2026}

\newpage
\appendix
\onecolumn
\section{Theoretical and Implementation Details of the AAPA Framework}

This appendix provides a deeper exploration of the theoretical underpinnings and implementation details of our proposed framework.

\subsection{The Unified View: SFT as a Special Case of Preference Learning}
\label{app:sft_as_pref}

Our framework treats SFT and RL as related post-training objectives that differ mainly in how supervision is expressed. To make this connection explicit, we reinterpret SFT as a limiting case of reward-based preference learning.

\paragraph{The Conventional View: SFT as Imitation Learning.}
Traditionally, SFT is understood as imitation learning. The objective is to train a policy $\pi_\theta$ to mimic an expert by minimizing the Negative Log-Likelihood (NLL) of a single ground-truth response $y^*$:
\begin{equation*}
\mathcal{L}_{\text{SFT}}(\theta) = -\log \pi_\theta(y^*|x)
\end{equation*}
Minimizing this loss is equivalent to maximizing the probability $\pi_\theta(y^*|x)$. In the conventional view, SFT therefore optimizes toward a provided target response, whereas preference learning usually compares or scores multiple candidate responses.

\paragraph{The Unified View: SFT as Preference Learning with a Dirac Delta Reward.}
Our framework connects these concepts by defining a degenerate reward model for the SFT case. While standard preference learning (e.g., DPO) is relative (a "winner" $y_w$ is preferred over a "loser" $y_l$), SFT can be interpreted as assigning all positive supervision to a single demonstrated response.

Specifically, the ground-truth response $y^*$ is treated as the unique supervised target, while other responses receive no direct SFT reward under this idealized view. To model this, we define a reward function using the Dirac delta function, $\delta(\cdot)$:
\begin{equation*}
R(y|x) = \delta(y - y^*)
\end{equation*}
This function assigns positive mass only when the policy generates the demonstrated response and zero mass to deviations.

\paragraph{Mathematical Equivalence of Reward Maximization and NLL Minimization.}
We can formally show that the standard SFT objective is a special case of the general reward-maximization objective.
\begin{enumerate}
    \item The standard reinforcement learning objective is to maximize the expected reward for a response $y$ sampled from the policy $\pi_\theta(\cdot|x)$:
    \begin{equation*}
    \max_{\theta} \mathbb{E}_{y \sim \pi_\theta(\cdot|x)} [R(y|x)]
    \end{equation*}
    
    \item Substituting our Dirac delta reward function:
    \begin{equation*}
    \max_{\theta} \mathbb{E}_{y \sim \pi_\theta(\cdot|x)} [\delta(y - y^*)]
    \end{equation*}
    
    \item The expectation is an integral (or sum) of the reward of each possible output $y$ weighted by its probability $\pi_\theta(y|x)$. Due to the properties of the Dirac delta function, the integral is non-zero only at the single point $y=y^*$. Therefore, the expectation collapses to be proportional to the probability of generating $y^*$:
    \begin{equation*}
    \mathbb{E}_{y \sim \pi_\theta(\cdot|x)} [\delta(y - y^*)] \propto \pi_\theta(y^*|x)
    \end{equation*}
    
    \item To maximize this expectation, one must maximize the probability term:
    \begin{equation*}
    \max_{\theta} \pi_\theta(y^*|x)
    \end{equation*}
    
    \item Since the logarithm is a monotonic function, this is equivalent to maximizing its logarithm:
    \begin{equation*}
    \max_{\theta} \log \pi_\theta(y^*|x)
    \end{equation*}
    
    \item Finally, maximizing a function is equivalent to minimizing its negation:
    \begin{equation*}
    \min_{\theta} (-\log \pi_\theta(y^*|x))
    \end{equation*}
\end{enumerate}
This final expression is precisely the NLL loss for SFT. Thus, SFT can be viewed as a special case of reward maximization in which the reward is concentrated on the expert demonstration.

\subsection{Theoretical Significance of the Unified View}
This view is useful for two reasons:
\begin{enumerate}
    \item \textbf{Connects Alignment Objectives.} It places SFT, DPO, and online RL under a common reward-maximization perspective, which helps explain why the same anchoring signal can be added to different post-training objectives.
    \item \textbf{Clarifies the Role of Supervision.} It frames post-training objectives as using reward signals with different levels of granularity. SFT concentrates supervision on a demonstrated target, while preference- or reward-based methods provide smoother feedback over multiple responses. AAPA builds on this view by adding a sentence-level anchor that supplies dense expert guidance during policy optimization.
\end{enumerate}

\subsection{POLAR-based Discriminator Implementation}
\label{app:adversarial_impl}

As described in the main text, our adversarial objective $\mathcal{L}_{\text{ADV}}$ is instantiated via a sentence-level discriminator $D_\phi$. In practice, any semantic discriminator operating at the sentence level can provide a meaningful adversarial signal. In this work, we adopt POLAR, a reference-based model originally designed to measure distances between generation strategies, as the backbone of our discriminator. The raw distance scores produced by POLAR are further transformed into a well-behaved adversarial loss coefficient compatible with our optimization objective. The complete transformation procedure is presented in Algorithm~\ref{alg:polar_to_coef}.

\begin{algorithm}
\caption{Adversarial Loss Coefficient Computation Based on POLAR Distance}
\label{alg:polar_to_coef}
\begin{algorithmic}[1]
\Require Prompt $x$, Teacher policy $\pi_t$, Student policy $\pi_s$
\Ensure Adversarial loss coefficient $coef \in [-1, 1]$
\State Obtain teacher response $y_t = \pi_t(x)$ and student response $y_s = \pi_s(x)$
\State Feed $(x, y_t, y_s)$ into POLAR model to get BT distance: $r = \text{POLAR}(x, y_t, y_s)$
\State Normalize distance $r$ into similarity score $p$ using sigmoid:
\Statex \(\displaystyle p = \sigma(r) = \frac{1}{1 + e^{-r}} \quad (p \in (0,1))\)
\State Convert similarity score $p$ into an adversarial loss coefficient $coef$:
\Statex \(\displaystyle coef = 1 - 8(p - 0.5)^2 \quad (coef \in [-1, 1])\)
\State \textbf{return} $coef$
\end{algorithmic}
\end{algorithm}
The final $coef$ serves as the output of our discriminator, $D_\phi(x, y_t, y_s)$, which is then used to compute $\mathcal{L}_{\text{ADV}}$.

\subsection{A-CHORD Formulation}
\label{app:achord_formulation}

CHORD combines an online policy-optimization term with an off-policy expert-imitation term. In our notation, its objective can be written as
\begin{equation*}
\mathcal{L}_{\text{CHORD}}(\theta)
= (1-\mu_t)\mathcal{L}_{\text{GRPO}}(\theta)
+ \mu_t \mathcal{L}_{\text{SFT}}^{\phi}(\theta),
\end{equation*}
where $\mathcal{L}_{\text{GRPO}}$ is computed from policy rollouts and verifiable rewards, $\mathcal{L}_{\text{SFT}}^{\phi}$ is the expert-trajectory imitation loss with optional token-level weighting, and $\mu_t$ denotes the dynamic global weight used to balance online exploration and expert imitation during training. The token-level weighting function $\phi$ controls the contribution of individual expert tokens, so the expert term mainly constrains uncertain or useful tokens rather than uniformly forcing the full expert trajectory.

A-CHORD keeps these original mechanisms unchanged and adds the AAPA sentence-level anchoring term:
\begin{equation*}
\mathcal{L}_{\text{A-CHORD}}(\theta; \phi)
= \mathcal{L}_{\text{CHORD}}(\theta)
+ \lambda_{\text{adv}}\mathcal{L}_{\text{ADV}}(\theta; \phi).
\end{equation*}
During each update, A-CHORD follows the same data flow as CHORD: policy rollouts are used for the online GRPO component, while expert responses are used for the imitation component. The only additional step is to compare the sampled policy response $y_s$ with the corresponding offline expert response $y_t$ using the fixed discriminator $D_\phi(x,y_t,y_s)$, and then use the resulting coefficient in $\mathcal{L}_{\text{ADV}}$. Thus, the adversarial term does not alter the reward function, the dynamic CHORD weight $\mu_t$, or the token-level expert weighting scheme. Instead, it supplies a sentence-level distributional anchor that complements CHORD's token-level expert imitation.

\subsection{Alternative Preference Optimization Formulations}
\label{app:pref_algos}

Our AAPA framework is compatible with a wide range of preference optimization algorithms. While the main text focuses on a policy-gradient approach (GRPO), here we detail prominent reward-model-free alternatives.

\subsubsection{Direct Preference Optimization (DPO)}
DPO is a widely-used algorithm that formulates preference learning as a direct classification problem on human preferences.

\paragraph{Data Format.} The preference dataset $D_{\text{PREF}}$ for DPO consists of tuples $(x, y_w, y_l)$, where for a given prompt $x$, $y_w$ is the "winner" (preferred) response and $y_l$ is the "loser" (dispreferred) response.

\paragraph{Loss Function.} The DPO loss is formulated to directly increase the likelihood of the winner response while decreasing the likelihood of the loser response, relative to a fixed reference policy $\pi_{\text{ref}}$. It is defined as:
\begin{equation*}
\mathcal{L}_{\text{DPO}}(\theta, \pi_{\text{ref}}) = -\mathbb{E}_{(x, y_w, y_l) \sim D_{\text{PREF}}} \left[ \log \sigma \left( \beta \log \frac{\pi_\theta(y_w|x)}{\pi_{\text{ref}}(y_w|x)} - \beta \log \frac{\pi_\theta(y_l|x)}{\pi_{\text{ref}}(y_l|x)} \right) \right]
\end{equation*}
Here, $\beta$ is a temperature parameter that controls the strength of the preference modeling, and $\sigma$ is the sigmoid function.

\subsubsection{Kahneman-Tversky Optimization (KTO)}
KTO offers a different perspective by removing the need for pairwise comparisons altogether. It instead relies on binary labels for individual examples.

\paragraph{Data Format.} The preference dataset $D_{\text{PREF}}$ for KTO consists of tuples $(x, y, l)$, where $l \in \{ \text{desirable}, \text{undesirable} \}$ is a binary label indicating whether the response $y$ to prompt $x$ is good or bad.

\paragraph{Objective.} The KTO loss is designed based on principles from human prospect theory. It has two main components: one term encourages the policy to increase the likelihood of "desirable" examples, and another, more heavily weighted term, strongly discourages the policy from generating "undesirable" examples. This asymmetry reflects the human tendency to be more sensitive to losses than to equivalent gains.

\section{Detailed Benchmark Results and Examples}
\subsection{Detailed experiment training setting}
\label{app:experimental_settings}
\noindent\textbf{General Setting.}
All experiments were conducted on eight H100 GPUs.
The learning rate was set to $1\times10^{-5}$ for the SFT stage, $1\times10^{-7}$ for the GRPO stage, and $1\times10^{-6}$ for the mixed-objective training stage, following the default configuration of CHORD.
After completing one epoch of SFT, we perform one epoch of RL.
Under our unified training framework, the same SFT and RL datasets are used.
RL is trained with 8 rollouts, and the ratio between SFT and RL samples at each training step is fixed at $1{:}4$.
Consequently, the SFT data are trained for a total of two epochs, while the RL data correspond to one epoch.
Early stopping is not applied to ensure fair and consistent comparison across different training settings.

\noindent\textbf{Chat Model Setting.}
Since the Qwen3 models require manual control of the \textit{think} process,
all GRPO experiments based on the Chat models were performed with \texttt{enable\_thinking=False}.
Each model response begins with the prefix \texttt{<think>\textbackslash n\textbackslash n</think>\textbackslash n\textbackslash n}.
Similarly, for SFT training on Chat models, this prefix was prepended to the response part of the data, and the loss on empty think segments was excluded from the computation.

\noindent\textbf{Base Model Setting.}
To validate the effectiveness of our approach on the base model and ground-truth references, we conducted several ablation studies on the Qwen3-Base model using the reference answers. To eliminate the potential interference of reasoning-related labels, we applied supervised fine-tuning (SFT) with the \texttt{Qwen2.5} chat template, and removed the \texttt{<think>\textbackslash n\textbackslash n</think>\textbackslash n\textbackslash n} prefix from the SFT response data. Since A-SFT requires an additional rollout step and the base model lacks inherent conversational ability, experiments based on the base model were carried out under the SFT\textrightarrow GRPO training paradigm. As shown in Table~\ref{tab:ablationchatbase}, when using the ground-truth (GT) reference responses from the dataset as evaluation targets, initializing from the base model achieves better generalization performance than initializing from a chat model, while reaching comparable instruction-following capability. 

\begin{table}[H]
    \renewcommand\arraystretch{1.1} 
    \centering
    \begin{tabular}{lcccccc}
        \hline
        \hline
        \toprule
        \multirow{2}{*}{\textbf{\textit{Method}}}  & \multicolumn{6}{c}{Benchmarks} \\
        \cmidrule(lr){2-7}
                & Avg & I-Following & English & Coding & Mathematics & Chinese  \\
        \midrule
        \textbf{\textit{0.6B-Basenothink/GT}}   & - & - & - & - & - & - \\  
        \textbf{\textit{SFT}}           & 43.97 & 60.13 & 33.32 & 35.87 & 43.60 & \textbf{52.43} \\
        \textbf{\textit{SFT$\rightarrow$ GRPO}}      & 47.53 & 76.25 & \textbf{36.43} & \textbf{36.29} & 43.83 & 52.22 \\
        \textbf{\textit{SFT$\rightarrow$ A-GRPO}}    & \textbf{47.83} & \textbf{78.96} & 36.23 & 35.57 & \textbf{43.91} & 52.26  \\
        \bottomrule
    \end{tabular}
    \caption{Experimental results of base model on the ground-truth (GT) dataset. SFT\textrightarrow GRPO corresponds to RLVR in Table~\ref{tab:ablationother}, while SFT and SFT\textrightarrow A-GRPO corresponds to the entry with rate 0.001 in Table~\ref{tab:ablationGT}.}
    \label{tab:ablationchatbase}
\end{table}

\subsection{Supplementary ablation analysis}
\label{app:supplyablation}
\noindent\textbf{Ablation on AAPA Discriminator Coefficients with GT References and KL Loss.}
We further conducted an ablation study on the AAPA discriminator coefficient using ground-truth (GT) reference answers. Specifically, the GT answers were treated as the target, and both the teacher and student outputs were passed through POLAR to obtain their respective scores, $r_t$ and $r_s$. In the policy space, this corresponds to the teacher and student policies lying within a circle centered at the reference policy with radius $\max(r_t, r_s)$. The difference between these scores, $\Delta r$, approximates the distance $R$ between the two policies. Subsequently, the distances between these scores were calculated according to the algorithm described in the Appendix~\ref{app:adversarial_impl} and used to derive the discriminator coefficient. Table~\ref{tab:ablation_GTrefKL} presents the results of this ablation study. 
Finally, we conducted ablation experiments on the KL loss to examine its effect on training stability and performance. When the KL loss was disabled, the model's policy optimization was guided solely by the PG loss and adversarial loss.
Through this experiment, the presence or absence of reference answers in the calculation of the discriminator coefficient has little impact on the final performance. Disabling the KL loss allows the model to achieve better performance on this specific task, but it may lead to a collapse of its general capabilities on other tasks.
\begin{table}[htbp]
    \renewcommand\arraystretch{1.1}
    \centering
    \begin{tabular}{lcccccc}
        \hline
        \hline
        \toprule
        \multirow{2}{*}{\textbf{\textit{Method}}}  & \multicolumn{6}{c}{Benchmarks} \\
        \cmidrule(lr){2-7}
                & Avg & I-Following & English & Coding & Mathematics & Chinese  \\
        \midrule
        \textbf{\textit{Directly Distinguish}} & - & - & - & - & - & - \\  
        \textbf{\textit{Use KL}}   & 45.86 & 77.24 & 33.29 & \textbf{37.41} & 38.84 & \textbf{52.31} \\
        \textbf{\textit{NO KL}}    & \textbf{47.52} & \textbf{80.65} & \textbf{36.03} & 35.39 & \textbf{41.90} & 52.21 \\
        \textbf{\textit{With GT Response}} & - & - & - & - & - & - \\  
        \textbf{\textit{Use KL}}   & \textbf{46.25} & 77.43 & 32.48 & 36.40 & 41.98 & \textbf{51.95} \\
        \textbf{\textit{NO KL}}    & 40.65 & \textbf{80.38} & \textbf{36.62} & \textbf{38.01} & 12.64 & 51.62 \\
        \bottomrule
    \end{tabular}
    \caption{Ablation study of AAPA discriminator coefficients with ground-truth references and KL loss. Detailed experimental results are provided in Table~\ref{tab:ablationother}.}
    \label{tab:ablation_GTrefKL}
    \vspace{-10pt}
\end{table}

\noindent\textbf{Additional Robustness Checks.}
\label{app:additional_robustness}
We further test AAPA under three robustness settings that complement the main ablations. First, Table~\ref{tab:gkd_distillation} compares AAPA with GKD, an on-policy distillation baseline that uses teacher feedback on self-generated responses, whereas AAPA adds a sentence-level anchoring signal during reward optimization. This comparison separates teacher-feedback distillation from the reward-guided anchoring used by AAPA. Second, Table~\ref{tab:seed_robustness} evaluates random-seed robustness and a direct discriminator-output variant in the 0.6B setting. Third, Table~\ref{tab:scale_robustness} evaluates whether the same anchoring idea remains effective when applied to 8B A-GRPO. Across these settings, AAPA consistently improves IFEval while preserving competitive grouped general-purpose performance.
\begin{table}[H]
    \renewcommand\arraystretch{1.1}
    \centering
    \begin{tabular}{lccccc}
        \toprule
        Method & IFEval & English & Coding & Mathematics & Chinese \\
        \midrule
        \textbf{\textit{SFT}} & 60.26 & 25.20 & 29.43 & 39.56 & \textbf{44.33} \\
        \textbf{\textit{A-SFT}} & 61.74 & 26.67 & \textbf{33.93} & 40.14 & 44.10 \\
        \textbf{\textit{GKD}} & 61.18 & 27.16 & 33.20 & 37.86 & 43.94 \\
        \textbf{\textit{A-GRPO}} & \textbf{82.26} & \textbf{31.36} & 19.79 & \textbf{41.04} & 43.96 \\
        \bottomrule
    \end{tabular}
    \caption{Comparison with on-policy distillation, reporting IFEval and grouped general-purpose metrics.}
    \label{tab:gkd_distillation}
\end{table}

\begin{table}[H]
    \renewcommand\arraystretch{1.1}
    \centering
    \begin{tabular}{lccccc}
        \toprule
        Method & IFEval & English & Coding & Mathematics & Chinese \\
        \midrule
        \textbf{\textit{Seed-88}} & 81.15 & \textbf{30.84} & 24.54 & 30.43 & \textbf{41.98} \\
        \textbf{\textit{Seed-88-straight}} & 78.93 & 26.94 & 19.71 & \textbf{35.89} & 34.28 \\
        \textbf{\textit{Seed-1234}} & \textbf{81.52} & 28.47 & \textbf{28.55} & 31.37 & 40.14 \\
        \bottomrule
    \end{tabular}
    \caption{Random-seed and discriminator-form robustness for the 0.6B setting.}
    \label{tab:seed_robustness}
\end{table}

\begin{table}[H]
    \renewcommand\arraystretch{1.1}
    \centering
    \begin{tabular}{lccccc}
        \toprule
        Method & IFEval & English & Coding & Mathematics & Chinese \\
        \midrule
        \textbf{\textit{Qwen3-8B}} & 82.26 & \textbf{61.96} & \textbf{83.72} & 61.11 & \textbf{81.79} \\
        \textbf{\textit{8B-GRPO}} & 87.62 & 57.98 & 73.13 & \textbf{70.56} & 76.54 \\
        \textbf{\textit{8B-A-GRPO}} & \textbf{89.65} & 59.56 & 75.18 & 65.45 & 77.11 \\
        \bottomrule
    \end{tabular}
    \caption{Scale robustness under the 8B A-GRPO setting.}
    \label{tab:scale_robustness}
\end{table}

\noindent\textbf{Ablation on AAPA Discriminator and Teacher.}
To test whether AAPA depends on a specific discriminator or teacher model, we conduct ablations on the discriminator backbone, teacher responses, and length-only anchoring. For discriminator robustness, we replace POLAR with \texttt{BGE-M3}, a text-similarity retrieval model, under the SFT$\rightarrow$A-GRPO paradigm. Table~\ref{tab:ablation_dis_and_teacher} shows that BGE-M3 also improves instruction-following performance, suggesting that AAPA does not rely on POLAR specifically as long as the discriminator provides a useful semantic signal.

For teacher robustness, we replace the default \texttt{Qwen3-235B-Instruct} teacher with a smaller \texttt{Qwen3-32B} teacher while keeping the student model fixed as \texttt{Qwen3-0.6B-Base}. Combined with the base-model results in Table~\ref{tab:ablationchatbase}, this suggests that AAPA is not overly sensitive to the exact teacher scale: the policy still explores through RL, while the adversarial signal provides guidance toward the expert response distribution.

For the length-only control, we replace semantic anchoring with response-length-based variants to test whether the gains can be explained merely by matching teacher response lengths. The detailed benchmark-level results in Table~\ref{tab:ablationother2} show that length matching alone does not consistently reproduce the gains of POLAR-based anchoring.
\begin{table}[H]
    \renewcommand\arraystretch{1.1}
    \centering
    \begin{tabular}{lccccc}
        \hline
        \hline
        \toprule
        \multirow{2}{*}{\textbf{\textit{Method}}}  & \multicolumn{5}{c}{Benchmarks} \\
        \cmidrule(lr){2-6}
                & IFEval & English & Coding & Mathematics & Chinese  \\
        \midrule
        \textbf{\textit{Discriminator Ablation}} & - & - & - & - & - \\  
        \textbf{\textit{BGE-M3}}   & \textbf{78.93} & \textbf{34.94} & 34.57 & \textbf{39.92} & 52.11 \\
        \textbf{\textit{POLAR}}    & 78.74 & 33.29 & \textbf{37.41} & 38.84 & \textbf{52.31} \\
        \textbf{\textit{Teacher Ablation}} & - & - & - & - & - \\  
        \textbf{\textit{Qwen3-235B-Instruct}}    & \textbf{78.74} & 33.29 & \textbf{37.41} & 38.84 & 52.31 \\
        \textbf{\textit{Qwen3-32B}}    & 77.82 & \textbf{35.15} & 36.70 & \textbf{39.26} & \textbf{52.38} \\
        \bottomrule
    \end{tabular}
    \vspace{4pt}
    \begin{tabular}{lccccc}
        \toprule
        \multirow{2}{*}{\textbf{\textit{Length-only Ablation}}} & \multicolumn{5}{c}{Benchmarks} \\
        \cmidrule(lr){2-6}
        & IFEval & English & Coding & Mathematics & Chinese \\
        \midrule
        \textbf{\textit{POLAR (4B)}} & 88.54 & \textbf{57.75} & \textbf{74.37} & 71.42 & 70.30 \\
        \textbf{\textit{Length-1}} & 88.17 & 53.42 & 73.05 & 70.16 & \textbf{71.16} \\
        \textbf{\textit{Length-2}} & \textbf{89.65} & 56.29 & 71.74 & \textbf{72.99} & 70.60 \\
        \bottomrule
    \end{tabular}
    \caption{Ablation study of AAPA discriminator, teacher responses, and length-only anchoring. We report IFEval and grouped general-purpose metrics, excluding aggregate Avg and I-Following columns. Detailed experimental results are provided in Table~\ref{tab:ablationother2}.}
    \label{tab:ablation_dis_and_teacher}
\end{table}

\noindent\textbf{Ablation on Different Tasks.}
To validate the broad effectiveness of our theory, we conducted experiments on both RLHF tasks involving reward model scoring and on mathematical tasks. For the reward model task, we used the same instruction data but removed the rule-based rewards, and employed the \texttt{internlm2-20b-Reward} model to score model responses. As shown in Table~\ref{tab:ablation_rm}, we report model performance at 1/3 and 2/3 of an epoch, since reward models are prone to reward hacking. Although the reward scores continue to increase during training, the performance of \texttt{Qwen3-0.6B-Chat} completely collapses when trained on 150K instruction data. However, with the addition of adversarial gradients, the degradation in instruction-following ability is  mitigated.

For the mathematical tasks, we employed the CHORD method and sampled 20K examples from OpenR1-Math-220k,  and the same subset was used for experiments with the Luffy  to ensure a fair comparison. To ensure model consistency, the ablation experiments for the mathematical tasks were conducted on \texttt{Qwen2.5-7B-Instruct}. Table~\ref{tab:ablation_math} presents the results after training for one epoch, where A-CHORD achieves the best performance.

\begin{table}[htbp]
    \renewcommand\arraystretch{1.1}
    \centering
    \begin{tabular}{lcccccc}
        \hline
        \hline
        \toprule
        \multirow{2}{*}{\textbf{\textit{Method}}}  & \multicolumn{6}{c}{Benchmarks} \\
        \cmidrule(lr){2-7}
                & Avg & I-Following & English & Coding & Mathematics & Chinese  \\
        \midrule
        \textbf{\textit{Internlm2-20b-Reward}} & - & - & - & - & - & - \\  
        \textbf{\textit{RM-1/3}}   & 27.14 & 42.03 & 19.62 & 28.58 & 25.39 & 24.71 \\
        \textbf{\textit{RM-2/3}}   & 4.54  & 16.65  & 1.25 & 0.92 & 0.50 & 7.05 \\
        \textbf{\textit{A-RM-1/3}} & 26.90 & 43.81 & 12.74 & 31.50 & 26.58 & 27.13 \\
        \textbf{\textit{A-RM-2/3}} & 6.44  & 19.62 &1.44 & 0.31 & 4.63 & 9.59 \\
        \bottomrule
    \end{tabular}
    \caption{Ablation study on the reward model task. Although this task is prone to reward hacking, the addition of adversarial gradients mitigates such behavior. Detailed experimental results are provided in Table~\ref{tab:ablationother2}.}
    \label{tab:ablation_rm}
\end{table}

\begin{table}[htbp]
    \renewcommand\arraystretch{1.1}
    \centering
    \begin{tabular}{lcccc}
        \hline
        \hline
        \toprule
        \multirow{2}{*}{\textbf{\textit{Method}}}  & \multicolumn{4}{c}{Benchmarks} \\
        \cmidrule(lr){2-5}
                & Avg & AIME24 & AIME25 & AMC  \\
        \midrule
        \textbf{\textit{Qwen2.5-7B-Instruct}} &20.72 & 11.7 & 6.66 & 43.80  \\  
        \textbf{\textit{Luffy}} &23.69& 10.00 & 14.16 & \textbf{46.92}  \\  
        \textbf{\textit{CHORD}}  &24.25& \textbf{16.25} & 15.00 & 41.51 \\
        \textbf{\textit{A-CHORD}} & \textbf{24.64} &15.42 & \textbf{16.25} & 42.26 \\
        \bottomrule
    \end{tabular}
    \caption{Ablation study on the Math task. Evaluation was conducted on the AIME24, AIME25, and AMC datasets.}
    \label{tab:ablation_math}
\end{table}

\subsection{Detailed benchmark results}
\label{app:detailresult}

\begin{table}[H]
    \renewcommand\arraystretch{1} 
    \centering
    \begin{tabular}{lcccccccc}
        \hline
        \hline
        \toprule
        Benchmark & 0.6B& SFT & GRPO & S\textrightarrow G & S\textrightarrow A-G &A-SFT & A-GRPO & A-S\textrightarrow A-G \\
        \midrule
        Avg       & 40.81 & 38.48 & 40.93 & 42.56 & 43.61 & 39.83 & 41.99 & 43.31\\
        IFEVAL    & 55.64 & 60.26 & 74.68 & 80.78 & 82.07 & 61.74 & 82.26 & 82.99\\
        MultiIF   & 57.60 & 59.68 & 73.50 & 74.01 & 77.06 & 59.70 & 76.89 & 76.73\\
        Mmlu      & 43.91 & 46.16 & 42.01 & 43.79 & 43.35 & 47.07 & 40.54 & 45.28\\
        Mmlu-pro  & 23.63 & 21.37 & 26.36 & 24.61 & 25.99 & 23.34 & 25.77 & 27.08\\
        GPQA      & 30.30 & 8.08  & 28.79 & 36.87 & 28.79 & 9.60 & 27.78  & 28.79\\
        Humaneval & 41.46 & 40.85 & 34.76 & 37.20 & 37.20 & 41.46 & 35.98 & 39.63\\
        Mbpp      & 23.80 & 18.00 & 0.40  & 6.00  & 20.80 & 26.40 & 3.60 & 27.80\\
        Gsm8k     & 61.18 & 59.21 & 58.38 & 62.02 & 61.26 & 58.07 & 60.27 & 42.91\\
        Math-500  & 49.06 & 43.16 & 46.96 & 43.68 & 43.00 & 44.84 & 48.48 & 44.62\\
        TheoremQA & 16.62 & 16.32 & 19.12 & 13.00 & 15.38 & 17.50 & 14.37 & 15.88\\
        Cmmlu     & 45.36 & 45.99 & 45.00 & 45.88 & 46.01 & 46.44 & 44.84 & 46.60\\
        Ceval     & 41.14 & 42.66 & 41.22 & 42.88 & 42.18 & 41.76 & 43.08 & 41.38\\
        \bottomrule
    \end{tabular}
    \caption{Detailed separate and sequential experimental results on Qwen3-0.6B}
    \label{tab:Detailedmainstage0.6B}
\end{table}

\begin{table}[H]
    \renewcommand\arraystretch{1} 
    \resizebox{\linewidth}{!}{%
    \begin{tabular}{lcccccccc}
        \hline
        \hline
        \toprule
        Benchmark & 4B & SFT & GRPO & S\textrightarrow G & S\textrightarrow A-G &A-SFT & A-GRPO & A-S\textrightarrow A-G \\
        \midrule
        Avg       & 62.52 & 60.80 & 68.11 & 65.75 & 66.10 & 60.67 & 68.39 & 67.93\\
        IFEVAL    & 78.19 & 77.82 & 87.06 & 88.91 & 90.02 & 81.15 & 88.54 & 90.02\\
        MultiIF   & 82.00 & 83.37 & 86.74 & 90.58 & 90.88 & 83.79 & 87.80 & 91.28\\
        Mmlu      & 69.90 & 72.78 & 70.02 & 71.47 & 70.96 & 71.91 & 70.00 & 69.86\\
        Mmlu-pro  & 35.56 & 40.51 & 54.88 & 53.43 & 54.59 & 31.29 & 51.46 & 56.35\\
        GPQA      & 30.81 & 10.10 & 35.35 & 22.22 & 19.19 & 13.13 & 41.41 & 27.27\\
        Humaneval & 75.61 & 76.83 & 78.66 & 79.88 & 82.32 & 79.27 & 81.10 & 81.71\\
        Mbpp      & 65.00 & 60.40 & 64.20 & 63.20 & 61.80 & 61.40 & 65.20 & 64.80\\
        Gsm8k     & 83.78 & 86.81 & 89.92 & 89.08 & 89.31 & 86.73 & 87.04 & 90.14\\
        Math-500  & 70.26 & 62.20 & 70.14 & 63.36 & 62.66 & 63.72 & 70.7 & 65.92\\
        TheoremQA & 24.38 & 21.75 & 39.50 & 25.75 & 29.62 & 19.50 & 37.62 & 34.88\\
        Cmmlu     & 72.09 & 72.18 & 72.03 & 72.48 & 72.63 & 71.80 & 72.07 & 72.45\\
        Ceval     & 62.62 & 64.84 & 68.80 & 68.58 & 69.22 & 64.34 & 67.70 & 70.45\\
        \bottomrule
    \end{tabular}
    }
    \caption{Detailed separate and sequential experimental results on Qwen3-4B}
    \label{tab:Detailedmainstage4B}
\end{table}

\begin{table}[H]
    \renewcommand\arraystretch{1} 
    \begin{tabularx}{\linewidth}{l *{10}{>{\centering\arraybackslash}X}}
        \hline
        \hline
        \toprule
        \multirow{2}{*}{Benchmark} & \multicolumn{2}{c}{AAPA Model} & \multicolumn{6}{c}{Qwen3 Model} \\
        \cmidrule(lr){2-3} \cmidrule(l){4-10}
                  & 0.6B  & 4B    & 0.6B  & 1.7B  & 4B    & 4B-2507 & 8B&32B &235B-2507 \\
        \midrule
        Avg       & 43.31 & 67.93 & 40.81 & 52.83 & 62.52 & 65.34 & 72.43 & 67.19 & 77.05\\
    \textbf{AvgIF(Report)}& \textbf{79.86} & \textbf{90.65} & \textbf{56.62} & \textbf{67.59} & \textbf{80.10} & \textbf{82.84} & \textbf{84.48} & \textbf{85.04} & \textbf{89.18}\\
        IFEVAL    & 82.99 & 90.02 & 55.64 & 66.73 & 78.19 & 80.59 & 82.26 & 84.29 & 87.99\\
        MultiIF   & 76.73 & 91.28 & 57.60 & 68.44 & 82.00 & 85.09 & 86.69 & 85.79 & 90.37\\
        Mmlu      & 45.28 & 69.86 & 43.91 & 59.05 & 69.90 & 75.89 & 82.78 & 73.89 & 88.39\\
        Mmlu-pro  & 27.08 & 56.35 & 23.63 & 20.53 & 35.56 & 33.10 & 63.20 & 20.13 & 39.45\\
        GPQA      & 28.79 & 27.27 & 30.30 & 24.24 & 30.81 & 25.76 & 39.90 & 45.45 & 59.60\\
        Humaneval & 39.63 & 81.71 & 41.46 & 67.07 & 75.61 & 83.54 & 89.63 & 87.20 & 96.95\\
        Mbpp      & 27.80 & 64.80 & 23.80 & 51.80 & 65.00 & 69.20 & 77.80 & 69.60 & 84.60\\
        Gsm8k     & 42.91 & 90.14 & 61.18 & 74.37 & 83.78 & 85.14 & 84.46 & 88.17 & 84.99\\
        Math-500  & 44.62 & 65.92 & 49.06 & 63.24 & 70.26 & 70.98 & 73.88 & 91.36 & 93.50\\
        TheoremQA & 15.88 & 34.88 & 16.62 & 20.50 & 24.38 & 21.25 & 25.00 & 14.75 & 18.12\\
        Cmmlu     & 46.60 & 72.45 & 45.36 & 59.96 & 72.09 & 76.76 & 84.87 & 73.64 & 89.44\\
        Ceval     & 41.38 & 70.45 & 41.14 & 58.02 & 62.62 & 76.76 & 78.70 & 72.02 & 91.21\\
        \bottomrule
    \end{tabularx}
    \caption{Detailed results of the reported experiments}
    \label{tab:Detailedmainonestage}
\end{table}

\begin{table}[H]
    \renewcommand\arraystretch{1} 
    \begin{tabularx}{\linewidth}{l *{6}{>{\centering\arraybackslash}X}}
        \hline
        \hline
        \toprule
        \multirow{2}{*}{Benchmark} & \multicolumn{3}{c}{0.6B Model} & \multicolumn{3}{c}{4B Model} \\
        \cmidrule(lr){2-4} \cmidrule(l){5-7}
                  & SFT & CHORD & ACHORD & SFT & CHORD & ACHORD \\
        \midrule
        Avg       & 40.81 & 41.19 & 42.09 & 62.52 & 69.30 & 69.27 \\
        IFEVAL    & 55.64 & 74.49 & 76.16 & 78.19 & 87.06 & 87.80 \\
        MultiIF   & 57.60 & 72.54 & 74.34 & 82.00 & 86.87 & 88.74 \\
        Mmlu      & 43.91 & 41.49 & 44.10 & 69.90 & 70.29 & 69.94 \\
        Mmlu-pro  & 23.63 & 25.53 & 25.33 & 35.56 & 55.90 & 56.04 \\
        GPQA      & 30.30 & 22.73 & 28.79 & 30.81 & 38.38 & 40.40 \\
        Humaneval & 41.46 & 35.98 & 35.98 & 75.61 & 81.71 & 77.44 \\
        Mbpp      & 23.80 & 16.20 & 16.20 & 65.00 & 65.80 & 63.20 \\
        Gsm8k     & 61.18 & 55.57 & 55.57 & 83.78 & 87.41 & 90.86 \\
        Math-500  & 49.06 & 43.86 & 43.86 & 70.26 & 68.34 & 72.30 \\
        TheoremQA & 16.62 & 16.12 & 16.12 & 24.38 & 47.25 & 41.25 \\
        Cmmlu     & 45.36 & 45.85 & 45.85 & 72.09 & 71.96 & 72.69 \\
        Ceval     & 41.14 & 43.94 & 43.94 & 62.62 & 70.63 & 70.62 \\
        \bottomrule
    \end{tabularx}
    \caption{Detailed results of the unified experiments}
    \label{tab:Detailedunified}
\end{table}

\begin{table}[H]
    \renewcommand\arraystretch{1} 
    \begin{tabularx}{\linewidth}{l *{6}{>{\centering\arraybackslash}X}}
        \hline
        \hline
        \toprule
        Benchmark & SFT & 0.1 & 0.01 & 0.001 & 0.0001 & 0.00001 \\
        \midrule
        Avg       & 43.97 & 38.18 & 47.53 & 47.83 & 47.67 & 47.61 \\
        IFEVAL    & 62.66 & 76.52 & 78.74 & 80.78 & 81.33 & 79.30 \\
        MultiIF   & 57.59 & 72.70 & 76.13 & 77.14 & 76.49 & 76.36 \\
        Mmlu      & 52.90 & 50.92 & 51.99 & 52.02 & 51.91 & 52.56 \\
        Mmlu-pro  & 27.88 & 24.89 & 28.19 & 27.38 & 27.42 & 26.72 \\
        GPQA      & 19.19 & 25.76 & 30.30 & 29.29 & 25.25 & 28.28 \\
        Humaneval & 32.93 & 3.66  & 35.37 & 32.93 & 34.15 & 33.54 \\
        Mbpp      & 38.80 & 35.80 & 38.20 & 38.20 & 38.60 & 37.00 \\
        Gsm8k     & 62.70 & 26.99 & 62.77 & 63.38 & 65.81 & 64.75 \\
        Math-500  & 45.10 & 16.44 & 42.22 & 43.46 & 44.00 & 44.54 \\
        TheoremQA & 23.00 & 21.61 & 22.50 & 24.88 & 22.75 & 23.75 \\
        Cmmlu     & 51.02 & 49.94 & 50.13 & 50.17 & 50.15 & 50.15 \\
        Ceval     & 53.84 & 52.94 & 53.87 & 54.34 & 54.21 & 54.21 \\
        \bottomrule
    \end{tabularx}
    \caption{Detailed experimental results on GT coef}
    \label{tab:ablationGT}
\end{table}
\begin{table}[H]
    \renewcommand\arraystretch{1} 
    \centering
    \begin{tabularx}{\linewidth}{l *{6}{>{\centering\arraybackslash}X}}
        \hline
        \hline
        \toprule
        Benchmark & SFT & 0.1 & 0.01 & 0.001 & 0.0001 & 0.00001 \\
        \midrule
        Avg       & 42.47 & 38.09 & 46.21 & 45.86 & 46.22 & 46.89 \\
        IFEVAL    & 57.12 & 75.97 & 78.93 & 78.74 & 78.74 & 78.19 \\
        MultiIF   & 59.53 & 73.09 & 76.49 & 75.70 & 75.42 & 75.62 \\
        Mmlu      & 52.82 & 49.81 & 52.82 & 53.09 & 53.17 & 53.47 \\
        Mmlu-pro  & 27.04 & 23.23 & 26.81 & 27.08 & 26.85 & 28.12 \\
        GPQA      & 10.10 & 28.28 & 30.30 & 19.70 & 27.27 & 24.75 \\
        Humaneval & 40.85 & 22.56 & 34.76 & 38.41 & 34.15 & 38.40 \\
        Mbpp      & 35.00 & 28.60 & 36.60 & 36.40 & 35.80 & 34.80 \\
        Gsm8k     & 61.64 & 21.23 & 62.47 & 63.53 & 63.31 & 64.59 \\
        Math-500  & 38.16 & 10.68 & 29.16 & 29.86 & 34.58 & 36.74 \\
        TheoremQA & 21.75 & 21.38 & 22.88 & 23.12 & 21.38 & 22.88 \\
        Cmmlu     & 50.91 & 49.93 & 50.66 & 50.51 & 50.81 & 50.83 \\
        Ceval     & 54.78 & 52.36 & 52.61 & 54.10 & 53.13 & 54.24 \\
        \bottomrule
    \end{tabularx}
    \caption{Detailed experimental results on Teacher coef}
    \label{tab:ablationteacher}
\end{table}

\begin{table}[H]
    \renewcommand\arraystretch{1} 
    \resizebox{\linewidth}{!}{%
    \begin{tabular}{lcccccccc}
        \hline
        \hline
        \toprule
        Benchmark & RLVR & POLAR+RLVR & POLAR+$\mathrm{A_{coef}}$ & w $\mathrm{A_{coef}}$ & NoKL & REF & REFNoKL  \\
        \midrule
        Avg       & 47.52 & 46.70 & 47.68 & 47.83 & 47.52 & 46.25 & 40.65 \\
        IFEVAL    & 76.52 & 69.87 & 74.86 & 80.78 & 83.55 & 79.11 & 82.26 \\
        MultiIF   & 72.70 & 67.06 & 71.32 & 77.14 & 77.74 & 75.74 & 78.50 \\
        Mmlu      & 52.60 & 52.78 & 53.14 & 52.02 & 52.91 & 53.26 & 51.84 \\
        Mmlu-pro  & 27.89 & 29.00 & 29.48 & 27.38 & 27.90 & 27.02 & 26.71 \\
        GPQA      & 28.79 & 26.26 & 25.25 & 29.29 & 27.27 & 17.17 & 31.31 \\
        Humaneval & 35.37 & 37.20 & 42.07 & 32.93 & 36.00 & 37.20 & 38.40 \\
        Mbpp      & 37.20 & 36.40 & 35.80 & 38.20 & 34.80 & 35.60 & 37.60 \\
        Gsm8k     & 64.29 & 66.72 & 66.03 & 63.38 & 63.91 & 65.05 & 7.20 \\
        Math-500  & 43.70 & 46.10 & 46.62 & 43.46 & 38.90 & 37.40 & 8.34 \\
        TheoremQA & 23.50 & 25.37 & 22.62 & 24.88 & 22.88 & 23.50 & 22.38 \\
        Cmmlu     & 50.67 & 50.48 & 50.48 & 50.17 & 50.21 & 50.58 & 50.21 \\
        Ceval     & 53.77 & 53.16 & 54.54 & 54.34 & 53.03 & 53.32 & 53.03 \\
        \bottomrule
    \end{tabular}
    }
    \caption{Detailed experimental results on other}
    \label{tab:ablationother}
\end{table}

\begin{table}[H]
    \renewcommand\arraystretch{1} 
    \resizebox{\linewidth}{!}{%
    \begin{tabular}{lccccccccc}
        \hline
        \hline
        \toprule
        Benchmark & BGE-M3 & Qwen3-32B & RM-1/3 & RM-2/3 & A-RM-1/3 & A-RM-2/3 & POLAR-4B & Length-1 & Length-2 \\
        \midrule
        Avg       & 46.00 & 46.17 & 27.14 & 4.54  & 26.90 & 6.44  & 72.48 & 71.19 & 72.25 \\
        IFEVAL    & 78.93 & 77.82 & 42.33 & 12.20 & 44.73 & 15.53  & 88.54 & 88.17 & 89.65 \\
        MultiIF   & 75.20 & 74.82 & 41.72 & 21.10 & 42.88 & 23.71  & - & - & - \\
        Mmlu      & 52.67 & 53.08 & 25.13 & 2.58  & 12.50 & 2.55  & 68.81 & 68.34 & 69.05 \\
        Mmlu-pro  & 26.39 & 27.09 & 6.47  & 0.17  & 2.49  & 0.77  & 56.46 & 54.56 & 56.38 \\
        GPQA      & 25.76 & 25.25 & 27.27 & 1.01  & 23.23 & 1.01  & 47.98 & 37.37 & 43.43 \\
        Humaneval & 32.93 & 37.20 & 22.56 & 1.83  & 24.39 & 0.61  & 83.54 & 81.10 & 79.27 \\
        Mbpp      & 36.20 & 36.20 & 34.60 & 0.00  & 38.60 & 0.00  & 65.20 & 65.00 & 64.20 \\
        Gsm8k     & 61.87 & 64.06 & 40.94 & 0.61  & 41.09 & 11.60 & 89.31 & 87.95 & 90.90 \\
        Math-500  & 33.38 & 31.10 & 26.86 & 0.28  & 27.66 & 0.68  & 82.34 & 81.78 & 82.44 \\
        TheoremQA & 24.50 & 22.62 & 8.38  & 0.62  & 11.00 & 1.62  & 42.62 & 40.75 & 45.62 \\
        Cmmlu     & 50.38 & 50.72 & 23.64 & 1.99  & 25.71 & 4.47  & 72.02 & 71.73 & 72.05 \\
        Ceval     & 53.84 & 54.04 & 25.78 & 12.10 & 28.54 & 14.70  & 68.58 & 70.59 & 69.14 \\
        \bottomrule
    \end{tabular}
    }
    \caption{Detailed experimental results for discriminator, teacher, reward-model, and length-only ablations.}
    \label{tab:ablationother2}
\end{table}

\begin{table}[H]
    \renewcommand\arraystretch{1.04}
    \resizebox{\linewidth}{!}{%
    \begin{tabular}{lcccccccc}
        \hline
        \hline
        \toprule
        Benchmark & A-SFT & GKD & Seed-88 & Seed-88-straight & Seed-1234 & Qwen3-8B & 8B-GRPO & 8B-A-GRPO \\
        \midrule
        MMLU & 47.07 & 43.12 & 40.93 & 31.06 & 35.90 & 82.78 & 74.88 & 74.80 \\
        MMLU-Pro & 23.34 & 22.71 & 22.30 & 22.99 & 20.72 & 63.20 & 59.67 & 56.90 \\
        GPQA & 9.60 & 15.66 & 29.29 & 26.77 & 28.79 & 39.90 & 39.39 & 46.97 \\
        HumanEval & 41.46 & 37.20 & 29.88 & 25.61 & 30.49 & 89.63 & 78.05 & 84.76 \\
        MBPP & 26.40 & 29.20 & 19.20 & 13.80 & 26.60 & 77.80 & 68.20 & 65.60 \\
        GSM8K & 58.07 & 55.19 & 39.50 & 51.71 & 43.97 & 84.46 & 91.36 & 91.74 \\
        MATH & 44.84 & 47.64 & 35.54 & 41.08 & 37.88 & 73.88 & 81.82 & 81.24 \\
        TheoremQA & 17.50 & 10.75 & 16.25 & 14.88 & 12.25 & 25.00 & 38.50 & 23.38 \\
        IFEval & 61.74 & 61.18 & 81.15 & 78.93 & 81.52 & 82.26 & 87.62 & 89.65 \\
        CMMLU & 46.44 & 44.87 & 43.33 & 31.84 & 42.74 & 84.87 & 76.38 & 76.60 \\
        CEval & 41.76 & 43.01 & 40.62 & 36.72 & 37.53 & 78.70 & 76.70 & 77.61 \\
        \bottomrule
    \end{tabular}
    }
    \caption{Detailed benchmark-level results for the additional robustness checks reported in Appendix~\ref{app:additional_robustness}.}
    \label{tab:additional_robustness_detail}
\end{table}

\vspace{4pt}

\vspace{20pt}
\subsection{Example of Training data}
\label{app:exampletrain}


\tcbset{
  myboxstyle/.style={
    colback=gray!10,        
    colframe=blue!20,         
    boxrule=0.8pt,          
    arc=2pt,                
    boxsep=2pt,
    left=4pt,right=4pt,     
    top=4pt,bottom=4pt,
    before skip=5pt,
    after skip=5pt,
    fontupper=\normalsize,
    fonttitle=\bfseries,    
    coltitle=black,
  }
}
\lstset{
    language=Python,
    basicstyle=\ttfamily\footnotesize,
    keywordstyle=\color{blue},
    stringstyle=\color{red},
    commentstyle=\color{green!50!black},
    showstringspaces=false,
    breaklines=true,
    tabsize=4
}

\vspace{20pt}
\begin{tcolorbox}[myboxstyle, title=Eva Func Example]
\begin{lstlisting}
class KeywordChecker(Instruction):
    def build_description(self, *, keywords=None):
        """Build the instruction description.
        Args:
            keywords: A sequence of strings representing the keywords that are expected in the response.
        Returns:
            A string representing the instruction description."""
        
        self.keywords = keywords
        self._description_pattern = "Include keywords {keywords} in the response."
        return self._description_pattern.format(keywords=keywords)
        
    def check_following(self, value):
        """Check if the response contain the expected keywords."""
        for keyword in self._keywords:
            if not re.search(keyword, value, flags=re.IGNORECASE):
                return False
        return True
\end{lstlisting}
\end{tcolorbox}

\clearpage

\begin{tcolorbox}[myboxstyle,breakable, title=Training Example]
\textbf{Question}\\
Jamal works at a library shelving books. He has a cart full of books to put away in different sections. In the history section, he shelves 12 books. In the fiction section, he shelves 19 books. In the children’s section, he shelves 8 books but finds 4 that were left in the wrong place that he adds to his cart to shelve elsewhere. He still has 16 books to shelve. How many books did he start with in the cart? Include a palindrome in your response. There should be 2 paragraphs. Paragraphs are separated by the markdown divider: \textbf{***}
Include keywords ['item', 'peak', 'population', 'team'] in the response. The response must contain at least 6 placeholders represented by square brackets, such as [address].
\\[4pt]

\textbf{Ref Answer}\\
Jamal started his day at the library, a place where the [population] of book lovers often gathers. His task was to shelve books that had been returned or misplaced. In the history section, he carefully placed 12 books on the shelves, ensuring each [item] was in its correct spot. Moving on to the fiction section, he shelved 19 more books, making sure that the novels were in alphabetical order. In the children’s section, he shelved 8 books but found 4 that were misplaced, which he added back to his cart. After all this, he still had 16 books to shelve. To find out how many books Jamal started with, we need to add the books he shelved (12 + 19 + 8) to the books he found (4) and the books he still had to shelve (16). This gives us a total of [total\_books] books.\\ \textbf{***}
Interestingly, as Jamal was shelving books, he noticed a unique book titled “Peak” by a lesser-known author. The book’s title was a [palindrome], meaning it reads the same backward as forward. This intrigued him, and he made a mental note to read it later. As he continued his work, he realized that the library [team] was like a well-oiled machine, with everyone contributing to the smooth operation of the facility. By the end of his shift, Jamal felt a sense of accomplishment, knowing that he had played a crucial role in maintaining the library’s organization. He started with [total\_books] books in his cart, ensuring that each [item] found its rightful place on the shelves.\\[4pt]

\textbf{Teacher Answer}\\
Jamal began his shift with a cart full of books, each book an item awaiting its proper place. He first visited the history section, where he shelved 12 books, followed by the fiction section, where he placed 19 more. In the children’s section, he managed to shelve 8 books but discovered 4 misplaced ones, which he added back to his cart for later organization. Despite these comings and goings, he still has 16 books left to shelve. To determine how many books he started with, we calculate: 12 (history) + 19 (fiction) + 8 (children’s) - 4 (added back) + 16 (remaining) = 51 books total. This number represents the peak effort required by Jamal’s team to maintain order across the library's vast population of literary works. Each shelving decision impacts the accessibility of every item, much like how data at [address] influences urban population trends.

\textbf{***}The library's system relies on consistent workflows, similar to how a well-coordinated team manages resources at [facility\_name]. The discovery of misplaced books highlights the importance of audits, especially during peak hours when the population of visitors surges. Jamal’s cart, once full, now holds only 16 books—an improvement, but not yet complete. The palindrome "121" reflects the symmetry needed in such organizational tasks, where every action has a balanced reaction. Additional considerations include [city\_name], [state\_name], [zip\_code], [phone\_number], [email], and [department\_head], all of which support the larger mission of knowledge distribution. Ultimately, each item shelved brings the library closer to optimal order, mirroring how accurate data at [address] supports demographic studies on population shifts.
\\[4pt]

\textbf{Instruction ID} \\
\textquotesingle keywords\textquotesingle :\textquotesingle palindrome\textquotesingle,
\textquotesingle paragraphs\textquotesingle :\textquotesingle paragraphs\textquotesingle,\\
\textquotesingle keywords\textquotesingle :\textquotesingle existence\textquotesingle, 
\textquotesingle detectable\_content\textquotesingle :\textquotesingle number\_placeholders\textquotesingle

\textbf{Kwargs} \\
 None, \{\}, \{\textquotesingle keywords\textquotesingle: ['item', 'peak', 'population', 'team'\}, \{'num\_placeholders': 6\}
\end{tcolorbox}

\clearpage

\subsection{Example of answer between different training paradigm}
\label{app:badcase}

\tcbset{
  myboxstyle/.style={
    colback=gray!10,        
    colframe=blue!20,         
    boxrule=0.8pt,          
    arc=2pt,                
    left=4pt,right=4pt,     
    top=4pt,bottom=4pt,
    fonttitle=\bfseries,    
    coltitle=black,
  }
}

\begin{tcolorbox}[myboxstyle,breakable, title=Response Example]
\textbf{Instruction:} \\
Please explain in detail the process of photosynthesis in plants, including its stages, key molecules, and overall importance. In your explanation, make sure to include the keywords \texttt{['chlorophyll', 'sunlight', 'glucose', 'oxygen']}.  \\[4pt]

\textbf{Teacher Answer (Good, Long, Correct)}  \\
Photosynthesis is a fundamental biological process by which green plants, algae, and certain bacteria convert light energy from the sun into chemical energy stored in glucose molecules. This process occurs primarily in the chloroplasts of plant cells, which contain the pigment chlorophyll that captures sunlight efficiently. Photosynthesis can be divided into two major stages: the light-dependent reactions and the Calvin cycle (light-independent reactions). During the light-dependent reactions, chlorophyll absorbs sunlight and drives the splitting of water molecules, releasing oxygen as a byproduct and generating energy carriers ATP and NADPH. In the Calvin cycle, carbon dioxide from the atmosphere is fixed into glucose using the ATP and NADPH generated in the first stage. This glucose serves as a primary energy source for the plant, while the oxygen released supports the respiration of other living organisms. Understanding this process is crucial for fields such as agriculture, ecology, and environmental science, as it underpins plant growth, food production, and global oxygen balance.  \\[4pt]

\textbf{Model Answer (Incorrect, Short)}  \\
Photosynthesis is how plants get energy from sunlight.  \\[4pt]

\textbf{SFT Model Answer (Incorrect, Long)}  \\
Photosynthesis occurs when plants absorb sunlight using chlorophyll and turn it into energy that they can use immediately. The process involves creating sugar in the leaves and distributing it throughout the plant for growth. While it generates glucose, the role of oxygen release and carbon fixation is misunderstood, making the explanation inaccurate.  \\[4pt]

\textbf{GRPO Answer (Correct, Short)}  \\
Plants convert sunlight, water, and carbon dioxide into glucose and oxygen using chlorophyll.  \\[4pt]

\textbf{Adversarial Answer (Correct, Medium-Long, Similar to Teacher)}  \\
Photosynthesis is the complex process through which green plants convert sunlight into chemical energy stored in glucose. Chlorophyll in the chloroplasts captures sunlight, which drives the splitting of water molecules, releasing oxygen as a byproduct and creating ATP and NADPH. During the Calvin cycle, carbon dioxide is fixed into glucose using these energy carriers. This process is vital not only for plant growth but also for maintaining atmospheric oxygen levels and supporting other life forms. The keywords 'chlorophyll', 'sunlight', 'glucose', and 'oxygen' are all included as requested, ensuring the explanation is complete and accurate.\\[4pt]
\end{tcolorbox}


\end{document}